\documentclass[runningheads]{llncs}

\usepackage{eccv}

\usepackage{eccvabbrv}

\usepackage{graphicx}
\usepackage{booktabs}

\usepackage[accsupp]{axessibility}  % Improves PDF readability for those with disabilities.

\usepackage{amsmath}
\usepackage{multirow}
\usepackage{makecell}
\usepackage{color}
\usepackage{mathtools}
\usepackage{booktabs}
\usepackage{comment}
\usepackage{ifthen}
\usepackage{colortbl}
\usepackage{float} 
\usepackage{pifont}
\usepackage{enumitem}
\usepackage{threeparttable}
\usepackage{relsize}
\usepackage[T1]{fontenc}
\usepackage[font=small,labelfont=bf,tableposition=top]{caption}
\usepackage{rotating}
\usepackage{svg}
\usepackage[colorlinks]{hyperref}

\usepackage{hyperref}

\usepackage{orcidlink}

\DeclareCaptionLabelFormat{andfigure}{#1~#2  \&  \figurename~\thefigure}

\def\FastDiMEnonMask{FastDiME (w/o Mask)}
\def\FastDiME{FastDiME}
\def\FastDiMETwo{FastDiME-2}
\def\FastDiMETwoPlus{FastDiME-2+}

\newcommand\blfootnote[1]{%
  \begingroup
  \renewcommand\thefootnote{}\footnotetext{#1}%
  \addtocounter{footnote}{-1}%
  \endgroup
}

\begin{document}

% ---------------------------------------------------------------
\title{Fast Diffusion-Based Counterfactuals for Shortcut Removal and Generation}

\titlerunning{Counterfactuals for Shortcut Learning Detection}

\author{Nina Weng\inst{1\ast}\orcidlink{0009-0006-4635-0438} \and
Paraskevas Pegios\inst{1,2\ast}\orcidlink{0009-0005-1471-4850} \and
Eike Petersen\inst{1}\orcidlink{0000-0003-0097-3868} \and
Aasa Feragen\inst{1,2}\orcidlink{0000-0002-9945-981X} \and
Siavash Bigdeli\inst{1}\orcidlink{0000-0003-2569-6473}}

\authorrunning{Weng and Pegios et al.}

\institute{Technical University of Denmark, Kongens Lyngby, Denmark\\
\email{\{ninwe,ppar,ewipe,afhar,sarbi\}@dtu.dk}\\
\and
Pioneer Centre for AI, Copenhagen, Denmark
}

\maketitle

\begin{abstract}
Shortcut learning is when a model -- e.g.~a cardiac disease classifier -- exploits correlations between the target label and a spurious shortcut feature, e.g.~a pacemaker, to predict the target label based on the shortcut rather than real discriminative features. This is common in medical imaging, where treatment and clinical annotations correlate with disease labels, making them easy shortcuts to predict disease.
We propose a novel detection and quantification of the impact of potential shortcut features via a fast diffusion-based counterfactual image generation that can synthetically remove or add shortcuts. Via a novel self-optimized masking scheme we spatially limit the changes made with no extra inference step, encouraging the removal of spatially constrained shortcut features while ensuring that the shortcut-free counterfactuals preserve their remaining image features to a high degree. Using these, we assess how shortcut features influence model predictions.

This is enabled by our second contribution: An efficient diffusion-based counterfactual explanation method with significant inference speed-up at comparable image quality as state-of-the-art. We confirm this on two large chest X-ray datasets, a skin lesion dataset, and CelebA.
Our code is publicly available at \href{https://fastdime.compute.dtu.dk/}{fastdime.compute.dtu.dk}.

  \keywords{Shortcut Learning \and Counterfactuals \and Diffusion Models}
\end{abstract}    
\section{Introduction}
\label{sec:intro}
\blfootnote{
${}^\ast$ N. Weng and P. Pegios contributed equally to this work.
}

\begin{figure}[t]
 \centering
  \includegraphics[width=0.97\linewidth]{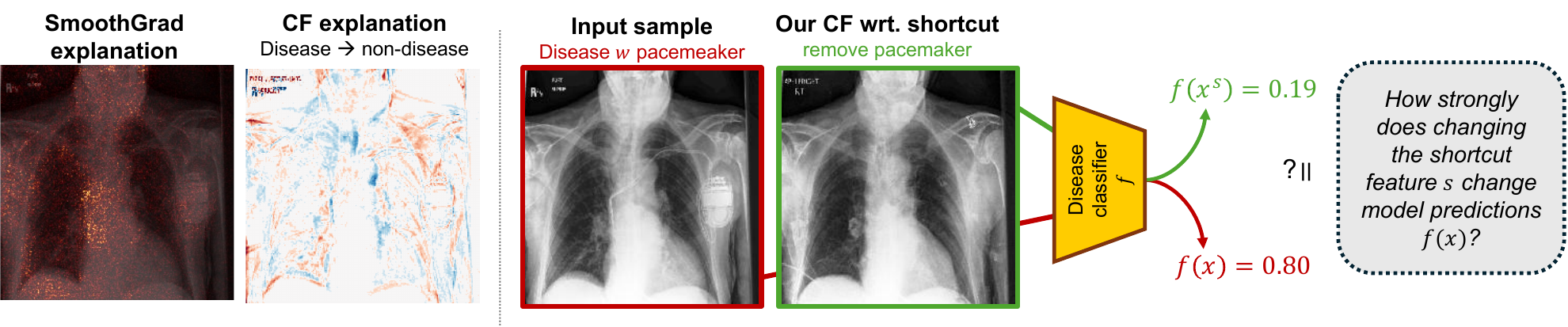}
  \caption{\textbf{Shortcut detection}: SmoothGrad and CF explanation (\textit{Left}), as two XAI methods, indicate which region in the image could influence the model decision (e.g. from disease to non-disease). Although this includes the shortcut features, it does not clearly indicate it. Therefore an expert is required for further visual inspection. Our counterfactual approach (\textit{Right}) only removes the desired shortcut attribute. With this we can validate if that specific attribute played a role in the model decision. 
  SmoothGrad visualization: highlighting areas crucial to the model decisions. CF explanation: difference map of the original image and CF (\textit{blue}/\textit{red}: information removal/addition).
}
  \label{fig:teaser}
\end{figure}

Shortcut learning~\cite{Geirhos2020} denotes the situation in which a model exploits a spurious correlation between a `shortcut' feature~$s$ and the target outcome~$y$. Known examples include grass being used as a shortcut for sheep, or surgical skin markers being shortcuts for malignant skin lesions~\cite{Winkler2019}.
Since the correlation is only spurious, and not causal, shortcut learning results in misleadingly high model performance on validation data that also contains shortcuts, but poor ability to generalize to data without shortcuts.
As prediction targets are often more complex than shortcuts, deep models can more easily and accurately identify the shortcuts in the data~\cite{Neuhaus2023, Jabbour2020, Gichoya2022}. Therefore, shortcut learning is a key obstacle to achieving well-generalized image classification models. This is especially true in medical imaging, where classification tasks often suffer from high intrinsic uncertainty and shortcuts are highly effective in improving classification performance~\cite{Winkler2019, OakdenRayner2020a, Badgeley2019, Jabbour2020, DeGrave2021, JimenezSanchez2023}.

Detecting shortcut learning is non-trivial.
Stratified analyses of model performance on samples with and without the potential shortcut might yield some indication, but cannot confirm shortcut learning as any performance gaps could result from other distributional differences. Explainable AI (XAI) methods have been proposed to obtain more concrete evidence~\cite{Zech2018, Sun2023, Thiagarajan2022, Pahde2023a}, but require manual inspection of explanations, complicating further use by explanation instability~\cite{Arun2021, Slack2021} and infinite sets of feasible competing explanations~\cite{Pawelczyk2020, Laugel2023} (see Fig.~\ref{fig:teaser}).
 
We propose a novel approach leveraging diffusion-based counterfactual (CF) generation~\cite{jeanneret2022diffusion, Augustin2022} to generate `shortcut counterfactuals', that \emph{do} or \emph{do not} contain a shortcut feature of interest.
Different from standard counterfactual explanations which change the target label, we rather seek to change the shortcut label. By assessing how predictions change with addition or removal of the shortcut feature, we quantify the model's degree of shortcut learning (see Fig.~\ref{fig:teaser}).

Our main contributions are the following:
\begin{enumerate}
    \item A diffusion-based method \FastDiME{} for counterfactual generation, employing approximate gradients in sampling, achieving a 20x speedup over comparable state-of-the-art models while maintaining counterfactual quality.
    \item A novel self-optimized masking scheme that confines counterfactual changes to a small region, with which we mitigate the unintentional removal of non-shortcut features in counterfactual images.
    \item A novel pipeline for detecting and quantifying shortcut learning via the generation of shortcut counterfactuals from our diffusion-based method.
    \item A demonstration of the generality of our method for counterfactual generation as well as its utility in detecting and quantifying shortcut learning in multiple realistic scenarios from different domains.
\end{enumerate}

\section{Related work}
\label{sec:related_work}

\paragraph{Counterfactual image explanations.}
A \emph{counterfactual} describes what the world, some outcome, or a given image would have looked like if some factor~$c$ would have had a different value.
The most prominent use of counterfactual inference in image analysis is in generating counterfactual explanations which represent images that are similar to a given image but would have been classified differently by a given image classification model.
In contrast to adversarial examples, not every minimally changed image that leads to a different classifier output represents a counterfactual example as this should also be realistic, i.e., close to the data manifold, and preserve the semantic properties of the original image.
Many methods have been proposed for visual counterfactual explanations, including VAEs~\cite{joshi2018xgems, Rodriguez2021}, GANs~\cite{Singla2020Explanation, lang2021explaining, nemirovsky2022countergan, jacob2022steex}, flow-based~\cite{dombrowski2021diffeomorphic} and diffusion-based models~\cite{Augustin2022,sanchez2022diffusion,jeanneret2022diffusion, jeanneret2023adversarial}.
In the medical imaging domain, different methods for generating healthy or diseased counterfactual images have been proposed~\cite{cohen2021gifsplanation, singla2023explaining, Thiagarajan2022, dravid2022medxgan, Mertes2022}. 

In contrast to most of the existing literature, our ultimate goal is \emph{not} only to generate counterfactual explanations of classifier decisions. Instead, we seek to change the state of a potential shortcut feature, such as a person's smile or the presence or absence of a cardiac pacemaker, from a given image, without changing the target class.
In this sense, our work is closely related to~\cite{pakzad2022circle, Pombo2023} which use StarGAN~\cite{Choi2018} for generating demographic counterfactuals in different medical image modalities, and to~\cite{DeSousaRibeiro2023} which combines a hierarchical VAE with structural causal models to perform principled counterfactual inference with respect to demographic attributes in brain MRI and chest X-ray images.
These methods target demographic attributes, however, we focus on other, often more localized potential shortcut features such as the presence or absence of cardiac pacemakers or chest drains. Notably, our method can serve dual purposes: a) as a counterfactual explanation method for explaining classifiers' decisions and b) as an image generation tool for our shortcut detection pipeline.

\paragraph{Diffusion-based counterfactuals.}
Denoising Diffusion Probabilistic Models~\cite{ho2020denoising} (DDPMs) have been successfully used for generating counterfactual explanations~\cite{Augustin2022,sanchez2022diffusion,jeanneret2022diffusion, jeanneret2023adversarial}. DiME~\cite{jeanneret2022diffusion} is the first method to adapt the original formulation of classifier guidance~\cite{dhariwal2021diffusion} for counterfactual generation but with a great increase in computational cost, since it requires back-propagation through the whole diffusion process to obtain gradients with respect to a noisy version of the input. DVCE~\cite{Augustin2022} introduces a cone-projection approach, assuming access to an adversarially robust copy of the classifier. Diff-SCM~\cite{sanchez2022diffusion} shares the encoder between the target model and the denoiser, making it model-specific. ACE~\cite{jeanneret2023adversarial} uses a DDPM to turn adversarial attacks into semantically meaningful counterfactuals. Recent studies~\cite{vaeth2023diffusion} report the high memory requirements and run time of diffusion-based methods as major challenges for large-scale evaluations. Inspired both by DiME~\cite{jeanneret2022diffusion} and ACE~\cite{jeanneret2023adversarial}, our method speeds up counterfactual generation and reduces memory usage significantly.

\paragraph{Guided diffusion and image editing.} 
Many approaches have been built to manipulate input images.
From these techniques, the most relevant works include Universal Guidance~\cite{bansal2023universal}, Motion Guidance~\cite{geng2024motion}, and GMD\cite{karunratanakul2023guided}. 
These methods improve the efficiency of the DiME approach by feeding the denoised image into the guiding classifier, and backpropagating the classifier gradients through the denoiser to calculate the gradients wrt. the input image.
We, instead, use the denoised image to approximate the gradients for the input image and avoid a heavy back propagation step.
The denoised image gradients act as a surrogate for the image gradients as they asymptotically reach the same image.

\paragraph{Shortcut learning detection.}
XAI-based methods provide explanations for individual decisions and thus may highlight the reliance of the model on potential shortcut features~\cite{Zech2018, DeGrave2021, Sun2023, Thiagarajan2022, Pahde2023a}.
While promising, these methods do not easily allow for quantitative analyses, require inspecting individual explanations, are limited to the detection of spatially localized shortcut features, and are subject to the general challenges of reliability and (non-)uniqueness~\cite{Arun2021, Slack2021, Pawelczyk2020, Laugel2023} making their use in helping users recognize the presence of shortcut learning limited~\cite{Adebayo2020, Adebayo2022}.
Thus, several methods have been proposed for finding potential shortcut features in a dataset~\cite{Zhang2023, Mueller2023, Neuhaus2023}, however, they do not evaluate whether a given model has indeed learned to exploit these shortcuts.
In~\cite{JimenezSanchez2023}, model performance is stratified by the presence or absence of a potential shortcut feature, finding large differences in average model performance between these groups. Yet, these performance variations might be influenced by other confounding factors. In~\cite{Brown2023}, models are retrained multiple times using a shortcut feature prediction head in a multi-task fashion, aiming to control and assess the degree to which the shortcut is encoded. Other approaches~\cite{Winkler2019,Nauta2021}, assess how adding surgical skin markings or colored patches to dermoscopic images affects model confidence. However, their shortcut features are simple and relatively easy to add or remove compared to, e.g., a cardiac pacemaker in a chest X-ray. To the best of our knowledge, no prior work has investigated the use of shortcut counterfactuals to quantify the degree to which a model relies on the shortcut feature.

\section{Methods}
\label{sec:method}

Counterfactual image explanation aims to solve the following problem: Assume given an image classification problem with a particular class $c$ of interest -- for instance, whether or not a chest X-ray contains a cardiac pacemaker -- and an image~$x$ that does not belong to class $c$. Can we provide an updated image~$x^c$ that remains as close as possible to~$x$ while both being visually realistic and undergoing sufficient visual change to clearly belong to the class~$c$? In our medical imaging case, this could consist of artificially adding or removing the pacemaker without changing any of the remaining patient anatomy. 

\subsection{Diffusion Models for Counterfactual Explanations (DiME)}

Jeanneret et al.~\cite{jeanneret2022diffusion} utilize DDPMs for generating counterfactual explanations through a guided diffusion process~\cite{dhariwal2021diffusion}. The image is guided towards the counterfactual class using the classifier's loss objective $L_{c}$, while the counterfactual $x^c$ is constrained to remain close to the original $x$ through an $L_1$ loss and a perceptual loss $L_{perc}$.
The overall gradient for the counterfactual loss term is of the form $\triangledown_{CF} = \lambda_{c}\triangledown L_c + \lambda_{1} \triangledown L_1 + \lambda_{p} \triangledown L_{perc}$, where $\lambda_{c}$, $\lambda_{1}$ and $\lambda_{p}$ are hyperparameters, and $L_1$ is measured between a noisy version and the original image.
To obtain meaningful gradients with their classifier, another nested diffusion process is used to synthesize (unconditionally) an image at the cost of running a whole DDPM process for each step.
DiME~\cite{jeanneret2022diffusion} is summarized as follows:

\begin{itemize}
\item Corrupt input up to noise level $\tau$ with the forward process.
\item For every time step  $ t \in \{\tau, \ldots, 0\}$, do:
\begin{enumerate}
    \item Denote noisy version of the counterfactual image $x^c_{t}$ .
    \item Compute the gradients $\triangledown_{CF}(\hat{x}^c_{t})$ based on a clean image $\hat{x}^c_{t}$ synthesized using an inner forward process continuing the unconditional generation process from the current step $t$, i.e, $\hat{x}^c_{t} = \text{DDPM(}x^c_{t}, t)$.
    \item Sample $x^c_{t-1}$ from $\mathcal{N}(\mu_g({x^c_t}),\Sigma({x^c_t}))$, where $\mu_g(x^c_t)$ is the guided mean, $\mu_g({x^c_t}) = \mu(x^c_t) - {\triangledown_{CF}} * \Sigma(x^c_t)$ and $\mu(x^c_t)$ is the estimated mean, following the standard guided diffusion scheme~\cite{dhariwal2021diffusion}.
\end{enumerate}
\item Return the counterfactual image as $x^c = x^c_{0}$.
\end{itemize}

In this process, step~2 is an expensive bottleneck. We therefore propose a modification that significantly reduces complexity and improves stability.

\begin{figure}[t]
  \centering
   \includegraphics[width=1.0\linewidth]{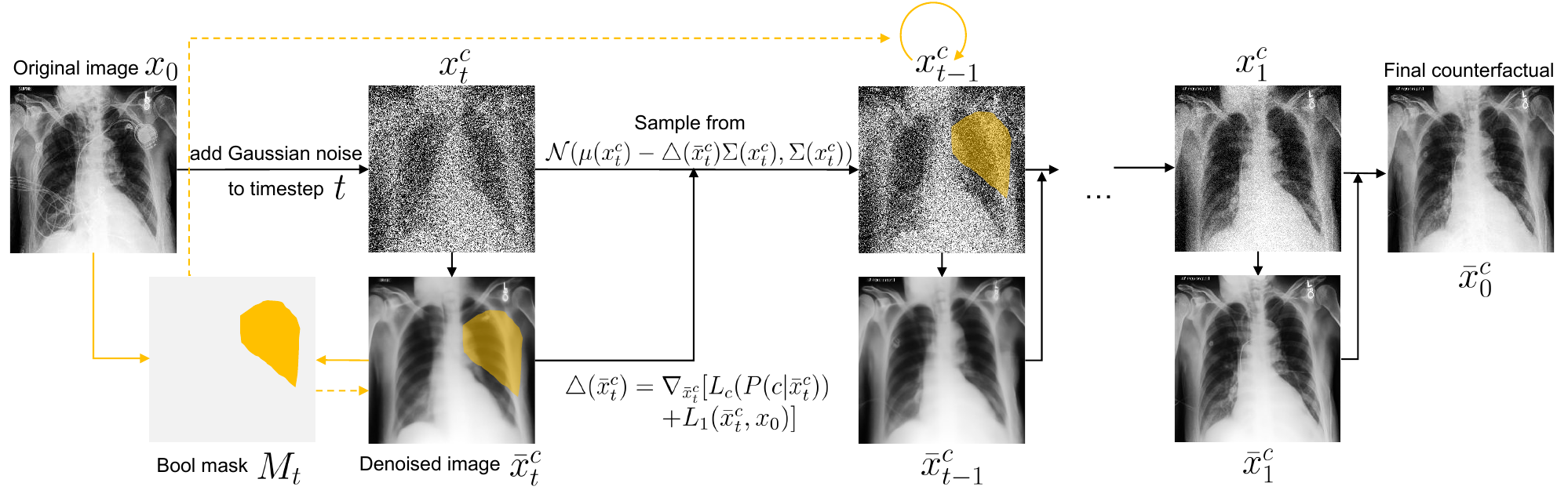}
   \caption{\textbf{Proposed \FastDiME{} method.} In each step, noised image $x_t^c$ is sampled with the guidance of the counterfactual loss, leveraging information derived from the denoised image $\bar{x}_t^c$. A self-optimized mask is automatically extracted and applied to prevent changes in regions less relevant to the task at each time step.}
   \label{fig:method_illustration}
\end{figure}

\subsection{Fast generation of high-quality counterfactuals}

We present \FastDiME{}, which improves on DiME with an efficient gradient estimation and a novel self-optimized masking scheme (see Fig.~\ref{fig:method_illustration}). %An illustration is shown in Fig.~\ref{fig:method_illustration}.
 
\paragraph{Efficient gradient estimation.}
As explained above, gradients in the original DiME model are retrieved from generated images by re-running the entire DDPM process. This is computationally expensive, with complexity $\mathcal{O}(T^2)$. To speed up inference while maintaining image quality, we propose using the denoised $\bar{x}^c_{t}$ to calculate the gradients.
In the forward diffusion process ($q$), the noisy image $x_t$ from $x_{t-1}$ is synthesized using a Gaussian distribution at each timestep with scheduled variance $\beta$: 
\begin{equation} \label{eq:noising_process}
q(x^c_t|x^c_{t-1}) \coloneqq \mathcal{N} (x^c_t; \sqrt{1-\beta_t} x^c_{t-1}, \beta_t \boldsymbol{I}).
\end{equation}

As derived by Ho et al.~\cite{ho2020denoising}, a direct  noising process conditioned on the input $x^c_0$ is feasible by marginalizing Eq.~\eqref{eq:noising_process},

\begin{equation} \label{eq:marginal_noising_process}
q(x^c_t|x^c_0) = \mathcal{N} (x^c_t; \sqrt{\bar{\alpha}_t}x^c_0, (1-\bar{\alpha}_t) \boldsymbol{I}), 
\end{equation}
\begin{equation} \label{eq:x_t_from_x_0}
x^c_t = \sqrt{\bar{\alpha}_t} x^c_0 + \sqrt{(1-\bar{\alpha}_t)} \epsilon,
\end{equation}

\noindent
where $\alpha_t \coloneqq 1 - \beta_t$, $\bar{\alpha}_t \coloneqq \prod_{s=0}^{t} \alpha_s$ and $\epsilon \sim \mathcal{N}(0,\boldsymbol{I})$.
Given a noise estimate~$\bar{\epsilon}$ (from a denoiser), we can use Eq.~\eqref{eq:x_t_from_x_0} to retrieve the estimated denoised $\bar{x}^c_t$ at time-step $t$ from the noisy image~$x^c_t$:
\begin{equation} \label{eq:x_0_from_x_t}
\bar{x}^c_{t} = \frac{x^c_t - \sqrt{(1-\bar{\alpha}_t}) \bar{\epsilon}}{\sqrt{\bar{\alpha}_t}}.
\end{equation}
By definition of the MSE-optimal denoiser \cite{raphan2011least, bigdeli2023learning}, we have
\begin{equation}
\bar{x}^c = \bar{x}^c_0 = \mathop{\mathbb{E}}[\hat{x}^c_{0}],
\end{equation}
which we use to calculate the gradients in counterfactual generation steps.
The gradients from $\bar{x}^c_{t}$ act as a surrogate for the image gradients $x^c_{t}$ as both images asymptotically reach the same solution in the diffusion process.
The time complexity of adopting a denoised image is $\mathcal{O}(T)$, which significantly expedites the entire inference procedure (refer to~\cref{sec:efficiencyanalysis} for time consumption experiments).

\begin{figure}[t]
\setlength{\tabcolsep}{2pt}
\centering
\begin{tabular}{cccc}
  \centering
   \includegraphics[width=0.23\linewidth]{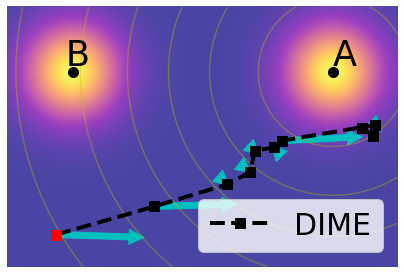}
   &
   \includegraphics[width=0.23\linewidth]{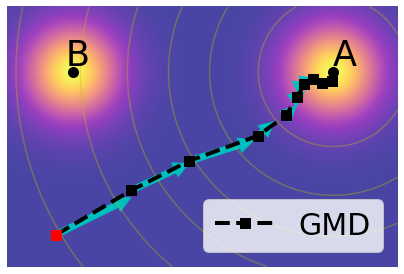}
   &
   \includegraphics[width=0.23\linewidth]{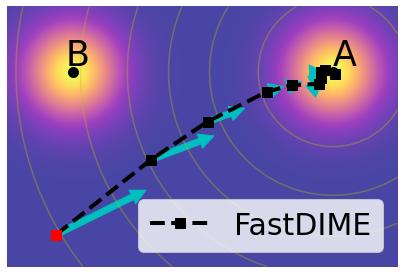}
   &
   \includegraphics[width=0.26\linewidth]{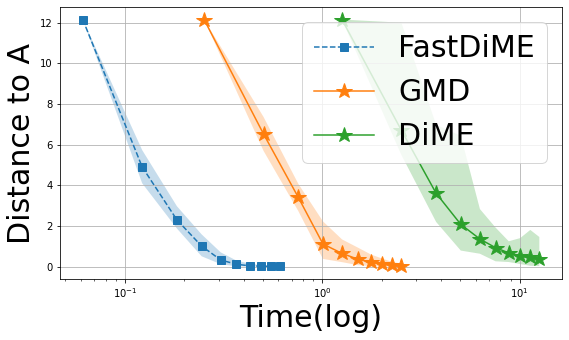}
\end{tabular}

   \caption{\textbf{Toy 2D example for visualization of the counterfactual generation convergence comparing DiME, GMD, and \FastDiME.}
   Given the two class distributions A and B, each method tries to bring the initial point $x_t$ (red point) to class A.
   Blue vectors indicate the gradients of classification loss $\triangledown L_c$ at each step.
   When calculating this gradient at each step, DiME uses a new unconditional sample from DDPM, which could lead to noisy gradients.
   In contrast, \FastDiME{} uses the expected image $\bar{x}_0$ at each step to calculate the gradients, which results in a more stable convergence.
   The plot on the right shows the convergence of each method in terms of the distance to A.
   This is averaged over 100 runs and it indicates that \FastDiME{} is accurate and significantly faster than both the other methods.}
   \label{fig:convergence}
\end{figure}

\cref{fig:convergence} illustrates the convergence of the proposed \FastDiME{} approach compared to DiME. In addition to the considerable advantage of improved time efficiency, using the denoised image does not degrade the generating quality, as the denoised image represents the expected value of all possible paths of generated images. Thus, the gradients computed on denoised images diminish noise levels and result in faster convergence with a more direct convergence route.

\paragraph{Self-optimized masking.}
Our application of counterfactual generation is to remove shortcut features from images. Shortcut features tend to be highly localized; thus, it is often desirable to constrain changes to the image made by the counterfactual generation process to a small region of the image.
To achieve this goal, Jeanneret et al.~\cite{jeanneret2023adversarial} perform an extra inpainting step after the usual counterfactual generation step using RePaint \cite{lugmayr2022repaint} to ensure highly localized changes to the image.
However, due to the in-painting step lacking classifier guidance, their approach does not guarantee that the generated images remain valid counterfactuals -- in principle, the class label could change again during the inpainting step.
Inspired by their approach, we integrate self-optimized masking directly into the initial counterfactual generation process to preserve more features from the original image, while incorporating other terms in the optimization. To this end, we first extract a mask $M_t$ by binarizing the difference between denoised image at time-step $t$ and original input $x_0$, and then masking the sampled image $x^c_t$ and denoised images $\bar{x}^c_{t}$ accordingly, i.e.,
\begin{align}
    M_t &= \delta(\bar{x}^c_{t},x_0)\\
    x^c_t{}' &= x^c_t \cdot M_t + x_t \cdot (1-M_t)\\
    \bar{x}^c_{t}{}' &= \bar{x}^c_{t}  \cdot M_t + x_0 \cdot(1- M_t)
\end{align}
where $x_t \sim q(x_t|x_0,t)$ refers to the sample from the forward process on the original image~$x_0$ that adds a Gaussian noise to it, and $\delta$ represents the function used for extracting mask. Note that the mask is not given, but automatically generated by our pipeline, and it can vary with each time-step $t$.
Different from \cite{jeanneret2023adversarial}, our fully gradient-guided masking approach ensures the validity of the counterfactuals. We implement this inside our main process after a warm-up period of $\tau_w$ time-steps, with $1 \leq \tau_w \leq \tau $. We further experiment with 2-step approaches keeping $M$ fixed after a completed run of guided diffusion with efficient gradient estimations or a completed run of our full method including our self-optimized mask scheme. We refer to these as \FastDiMETwo{} and \FastDiMETwoPlus{}, respectively.

\subsection{Shortcut learning detection} \label{sec:method_shortcut}

\begin{figure*}[t]
  \centering
    \includegraphics[width=1.0\linewidth]{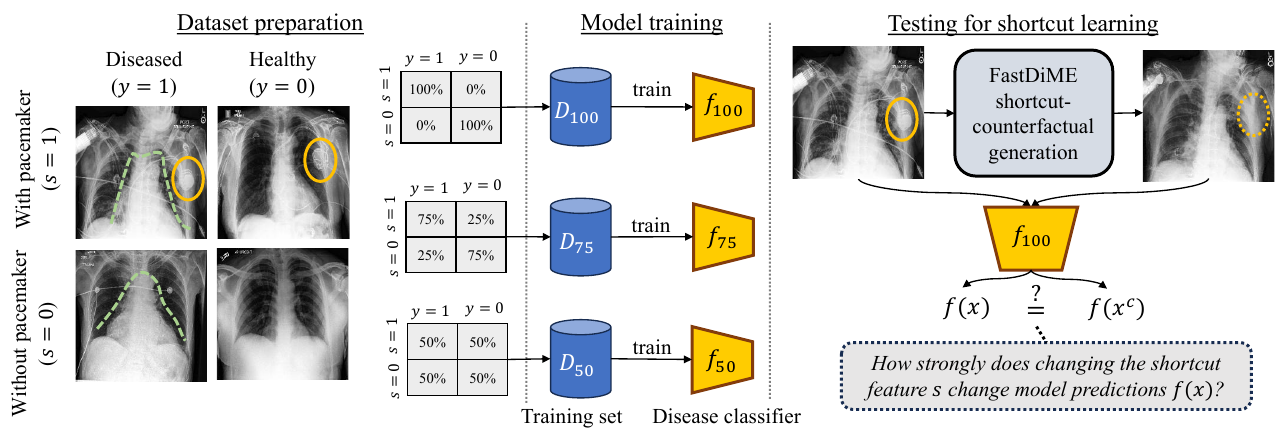}

   \caption{\textbf{Validating shortcut learning detection pipeline.} In order to test the shortcut detecting ability by shortcut-counterfactuals, we construct three synthetic training datasets with varying degrees of correlation between the shortcut feature~$s$ and the target label~$y$ and train classifiers based on them (\textit{Left } and \textit{Middle}). By measuring the difference in confidence leveling between the original image and shortcut-counterfactual, the degree of shortcut learning is examined (\textit{Right}).
   If model predictions differ strongly between natural images and shortcut counterfactual images (while leaving the target label unchanged), shortcut learning has occurred.}
   \label{fig:shortcut}
\end{figure*}

Assume a classifier $f(x) = \hat{y}$ that we suspect of 
shortcut learning, different from the one used for counterfactual image generation. 
For example, $f$ trained to diagnose lung diseases from chest X-ray images but instead might use medical equipment such as cardiac pacemakers or chest drains as shortcuts. We propose the pipeline for quantifying the degree to which $f$ has indeed learned to rely on the shortcut feature as shown in \cref{fig:shortcut} (\textit{Right}), where we generate the shortcut counterfactuals of all images, i.e. images that remove the pacemaker, and evaluate the alteration of predictions. 

In order to validate the proposed counterfactual-based shortcut detection pipeline, we propose the following framework (see \cref{fig:shortcut}):
\begin{enumerate}
\item Curate training sets with increasing levels of encoded correlation between the shortcut feature~$s=1$ and the target label~$y=1$, and train models on the increasingly contaminated training sets. In our experiments, we curate three training sets $\mathcal{D}_{k} , k=\{100,75,50\}$, in which $k\%$ of $y=1$ samples are also positive in $s$. It is worth noting that $\mathcal{D}_{100}$ is extremely biased by shortcut while $\mathcal{D}_{50}$ is completely balanced.
\item Curate natural test sets, along with a test set where the suspected shortcut feature is counterfactually flipped using \FastDiME{}. In particular, curate

\begin{enumerate}[label=(\alph*)] 
    \item an i.i.d. test set $\mathrm{test}_{k}$ with the same correlation rate as the training set,
    \item a balanced test set $\mathrm{test}_u$ 
    where each class (shortcut and target label) is represented equally with 50\% positive/negative samples,
    \item a counterfactual balanced test set called $\mathrm{test}^c_u$, which is obtained by generating shortcut counterfactuals~$x^c$ for each $x$ in the $\mathrm{test}_u$.
\end{enumerate}

\item Measure the difference in 
confidence level between original and shortcut-flipped counterfactual images.

\end{enumerate}
If the effect of the shortcut flipping operation on a classification model $f$ is stronger for the models trained on more biased datasets, this is a clear indication that the models are, indeed, prone to learning the suspected shortcut.

\section{Experiments and results} 
\label{sec:exp}

\subsection{Datasets and implementation details}

To ease comparison with existing work on counterfactual explanations~\cite{joshi2018xgems,rodriguez2021beyond,jacob2022steex,Augustin2022,jeanneret2022diffusion,jeanneret2023adversarial}, we evaluate  on \textbf{CelebA}~\cite{liu2015faceattributes}, containing 128 $\times$ 128 images of faces. First, we compare our counterfactual quality against state-of-the-art methods on the standard `smile' and `age' benchmarks.  To assess our shortcut detection pipeline, we set `smile' as a shortcut in predicting `age'. Furthermore, we validate our method on three real-world, challenging medical datasets. \textbf{CheXpert}~\cite{Irvin2019a} and \textbf{NIH}~\cite{Wang2017} are chest X-ray datasets with two different suspected shortcuts annotated; cardiac pacemakers~\cite{Jabbour2020} and chest drains~\cite{OakdenRayner2020a, JimenezSanchez2023}, respectively. Note that `pacemaker' and `chest drain' are potential shortcuts to chest X-ray diagnosis as they are common treatments for many cardio disease and lung diseases, respectively. \textbf{ISIC~2018}~\cite{Tschandl2018, Codella2019} is a skin lesion dataset with `ruler markers' that are reported to act as shortcuts for diagnosing whether lesions are malignant or not. For all medical datasets, images are resized to $224 \times 224$,  and we evaluate our method on shortcut counterfactuals, while diagnostic labels are used in shortcut detection pipeline experiments.

\paragraph{Implementation details.} For CelebA, we use the same trained DenseNet121 classifier~\cite{huang2017densely} and DDPM~\cite{ho2020denoising} as in DiME~\cite{jeanneret2022diffusion} and ACE~\cite{jeanneret2023adversarial} for fair comparisons. For all variants of our method, namely, \FastDiME, \FastDiMETwo, and \FastDiMETwoPlus, we follow the same hyperparameters as in DiME~\cite{jeanneret2022diffusion}. 
For the medical datasets, we train DDPMs with 1000 steps, use UNet's~\cite{ronneberger2015u} encoder architecture as the shortcut classifier for counterfactual generation, set $\tau$ to 160 out of 400 re-spaced time steps, and compute $L_{1}$ between the denoised and original input without including $L_{perc}$. 
For all datasets, we set $\tau_w = \frac{\tau}{2}$, 
and normalize, threshold and dilate the masks similar to ACE~\cite{jeanneret2023adversarial}.
For the shortcut detection experiments, the task classifier suspected of shortcut learning is a ResNet-18~\cite{he2016deep}.

\subsection{Counterfactual explanations}

\paragraph{Evaluation criteria.}
We follow the evaluation protocol of  ACE~\cite{jeanneret2023adversarial} on CelebA~\cite{liu2015faceattributes}. To measure \emph{realism} we use FID~\cite{heusel2017gans} (Fréchet Inception Distance) between the original images and their valid counterfactuals as well as sFID proposed in~\cite{jeanneret2023adversarial} to remove potential biases of FID. We estimate \emph{sparsity} using standard metrics for face attributes such as Mean Number of Attributes Changed (MNAC) which utilizes an oracle pre-trained on VGGFace2~\cite{cao2018vggface2} finetuned on CelebA~\cite{liu2015faceattributes}, as well as Correlation Difference (CD) proposed in~\cite{jeanneret2022diffusion} to account for MNAC's limitations. Moreover, we use Face Verification Accuracy (FVA)~\cite{cao2018vggface2} and Face Similarity (FS)~\cite{jeanneret2023adversarial} to measure whether a counterfactual changed the face identity. We also compute the \emph{transition probabilities} between the original image and its counterfactual with the COUnterfactual Transition (COUT) metric~\cite{khorram2022cycle}. Moreover, we include Bounded remapping of KL divergence (BKL), used in~\cite{jeanneret2022diffusion} to calculate the \emph{similarity} between prediction and the desired one-hot counterfactual label, with lower values indicating higher similarity. To measure the \emph{validity} of counterfactuals we report Flip Ratio (FR), i.e., the frequency of counterfactuals classified as the target label, and Mean Absolute Difference (MAD) of confidence prediction between original and counterfactual images. For medical datasets, we evaluate the quality of counterfactuals in \emph{closeness} and \emph{realism} using $L_1$ distance and FID respectively, and \emph{validity} using MAD.

\begin{table*}[t]
    \centering
    \caption{\textbf{Results for CelebA.} Results for DiVE~\cite{Rodriguez2021}, DiVE$^{100}$, STEEX~\cite{jacob2022steex}, DiME~\cite{jeanneret2022diffusion} and ACE~\cite{jeanneret2023adversarial} are from~\cite{jeanneret2023adversarial}. We compute BKL, MAD and FR metrics for diffusion-based methods, and highlight the \textbf{best} and \textit{second-best} performances.}
    \label{tab:celeba-main}
    \footnotesize
    
\scalebox{0.6}{\begin{tabular}{c|cccccccccc|cccccccccc} \toprule
        \multicolumn{1}{c}{} & \multicolumn{10}{|c|}{\textbf{Smile}} & \multicolumn{10}{c}{\textbf{Age}} \\\midrule
        Method        & FID  & sFID & FVA  & FS & MNAC  & CD  & COUT & BKL & MAD & FR & FID  & sFID & FVA  & FS & MNAC  & CD  & COUT & BKL & MAD & FR \\ \midrule
        DiVE          & 29.4 & -    & 97.3 & -  & -     & -    & -   & - & - & - & 33.8 & -    & 98.1 & -         & 4.58 & -  & - & - & - & -\\ 
        DiVE$^{100}$  & 36.8 & -    & 73.4 & -  & 4.63  & 2.34 & -   & - & - & - & 39.9 & -    & 52.2 & -  & 4.27 & - & - & - & - & - \\ 
        STEEX     & 10.2 & -    & 96.9 & -  & 4.11  & -    & -   & - & - & - & 11.8 & -    & 97.5 & -  & 3.44 & -  & - & - & - & -\\\midrule
         ACE $\ell_1$  & \textbf{1.27} & \textbf{3.97} & \textbf{99.9} & \textbf{0.874}     & 2.94  & \textit{1.73} & \textbf{0.783} & 0.199 & 0.72 & 0.976 & \textbf{1.45} & \textbf{4.12} & \textit{99.6} & \textit{0.782}    & 3.20 & \textit{2.94} & \textbf{0.718} & 0.266 & 0.60 & 0.962 \\
        ACE $\ell_2$  & \textit{1.90} & \textit{4.56} & \textbf{99.9} & \textit{0.867}     & \textbf{2.77}  & \textbf{1.56} & \textit{0.624}  & 0.326 & 0.59 & 0.843 & \textit{2.08} & \textit{4.62} & \textit{99.6} & \textbf{0.797}    & 2.94 & \textbf{2.82} & \textit{0.564} & 0.390 & 0.48 & 0.775\\
        DiME          & 3.17 & 4.89 & 98.3 & 0.729     & 3.72  & 2.30 & 0.526 & 0.094 & \textit{0.82} & 0.972 & 4.15 & 5.89 & 95.3 & 0.671    & 3.13 & 3.27 & 0.444 & 0.162 & \textbf{0.71} & 0.990 \\\midrule

        \FastDiME & 4.18 & 6.13 & \textit{99.8} & 0.758 & 3.12  & 1.91 & 0.445 & \textit{0.097} & \textit{0.82} & \textit{0.990} & 4.82 & 6.76 & 99.2 & 0.738   & 2.65 &  3.80 & 0.356 & 0.181 & \textit{0.69} & 0.986 \\
        
        \FastDiMETwo  & 3.33 & 5.49 & \textbf{99.9} & 0.773 & 3.06 & 1.89 & 0.439 & \textbf{0.083} & \textbf{0.83} & \textbf{0.994}  & 4.04 & 6.01 & \textit{99.6} & 0.750 & \textit{2.63} & 3.80 & 0.369 & \textbf{0.157} & \textbf{0.71} & \textbf{0.993} \\
        \FastDiMETwoPlus  & 3.24 & 5.23 & \textbf{99.9} & 0.785 & \textit{2.91} & 2.02 & 0.411 & 0.098 & \textit{0.82} & 0.989  & 3.60 & 5.59 & \textbf{99.7} & 0.766 & \textbf{2.44} & 3.76 & 0.323 & 0.179 & \textit{0.69} & \textit{0.987}
         
         \\\bottomrule 
    \end{tabular}}
    
\end{table*}

\begin{table}[t]
    \centering
    \caption{\textbf{Results for medical datasets.} We highlight the \textbf{best} and \textit{second-best} performances.}
\label{tab:medical_main}
    \small
    \scalebox{0.75}{\begin{tabular}{c|ccc|ccc|ccc}
    \toprule
    & \multicolumn{3}{c}{\textbf{CheXpert}} & \multicolumn{3}{|c|}{\textbf{NIH}} & \multicolumn{3}{c}{\textbf{ISIC}} \\
    \midrule
     Method & $L_1$ & MAD & FID & $L_1$ & MAD & FID& $L_1$ & MAD & FID \\
    \midrule
    DiME  & 0.1107&\textbf{0.7977}&73.2588 
                    & 0.0748 &\textbf{0.2536}& 94.3023
                    &0.0779&\textbf{0.5240}&121.5135\\
     \FastDiME &\textit{0.0897} &0.7554 &\textbf{61.4010}
                    & \textit{0.0546} &0.2244& \textit{64.1100}
                    &\textit{0.0631} &0.3732&\textbf{86.0475}\\
     \FastDiMETwo & 0.0946 & \textit{0.7596} & 64.0955 
                    & 0.0584 & \textit{0.2386} & 67.8026
                    & 0.0661 &  \textit{0.4428} & 87.4575\\
     \FastDiMETwoPlus & \textbf{0.0894} & 0.7581 & \textit{62.2840} 
                    & \textbf{0.0536} & 0.2263 & \textbf{63.2997}
                    & \textbf{ 0.0621} & 0.3905 &\textit{ 86.9823}\\
    \bottomrule

    \end{tabular}}

\end{table}

\begin{figure}[t]
\setlength{\tabcolsep}{1pt}
\centering
\begin{tabular}{cccc|ccc|ccc}
&
\multicolumn{3}{c}{ \scalebox{0.55}{\textbf{(a) CheXpert}: Chest X-rays with `pacemaker'. }} &
\multicolumn{3}{c}{ \scalebox{0.55}{\textbf{(b) NIH}: Chest X-rays with `chest drain'.} } &
\multicolumn{3}{c}{ \scalebox{0.55}{\textbf{(c) ISIC 2018}: Skin lesions with `ruler markers'.} }\\
\raisebox{7pt}{ \scalebox{.5}{\begin{sideways}{\footnotesize Original}\end{sideways}}} &
\includegraphics[width=0.1\linewidth]{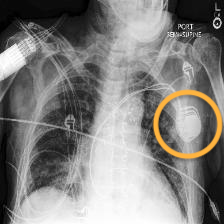} & \includegraphics[width=0.1\linewidth]{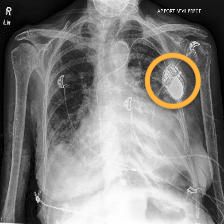} & \includegraphics[width=0.1\linewidth]{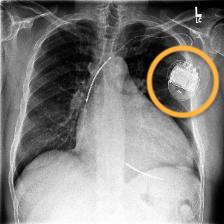} & 
\includegraphics[width=0.1\linewidth]{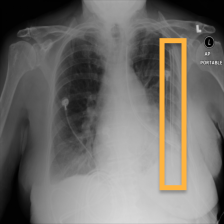} & 
\includegraphics[width=0.1\linewidth]{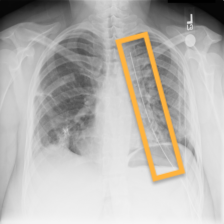} & 
\includegraphics[width=0.1\linewidth]{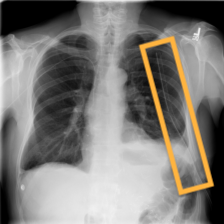} & 
\includegraphics[width=0.1\linewidth]{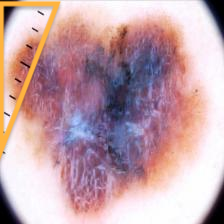} & 
\includegraphics[width=0.1\linewidth]{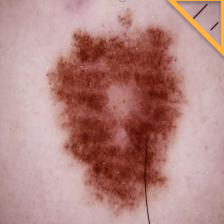} & 
\includegraphics[width=0.1\linewidth]{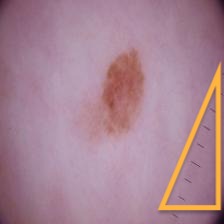} 
\\
\raisebox{7pt}{ \scalebox{.5}{\begin{sideways}{DiME}\end{sideways}} }&
\includegraphics[width=0.1\linewidth]{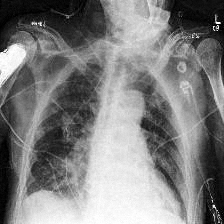}        & 
\includegraphics[width=0.1\linewidth]{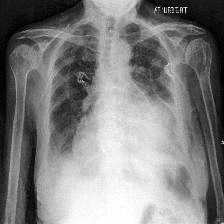}        & 
\includegraphics[width=0.1\linewidth]{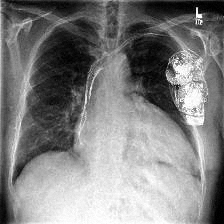}        &
\includegraphics[width=0.1\linewidth]{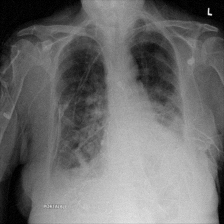}        & 
\includegraphics[width=0.1\linewidth]{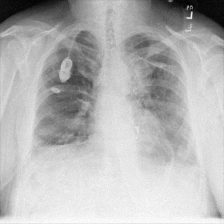}        & 
\includegraphics[width=0.1\linewidth]{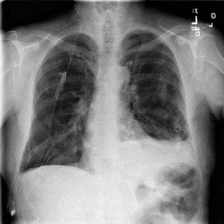}       &
\includegraphics[width=0.1\linewidth]{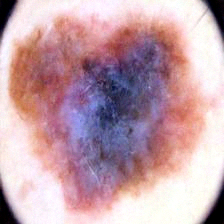}        & 
\includegraphics[width=0.1\linewidth]{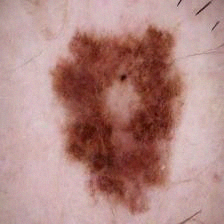}        & 
\includegraphics[width=0.1\linewidth]{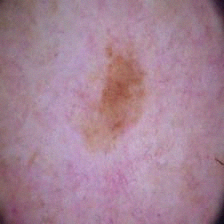}
\\
\raisebox{5pt}{ \scalebox{.5}{\begin{sideways}{\FastDiME}\end{sideways}} }&
\includegraphics[width=0.1\linewidth]{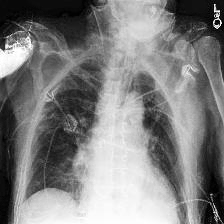}    & 
\includegraphics[width=0.1\linewidth]{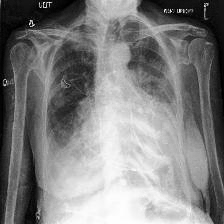}    & 
\includegraphics[width=0.1\linewidth]{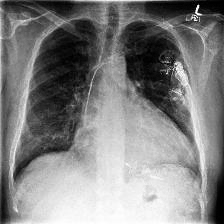}    &                           
\includegraphics[width=0.1\linewidth]{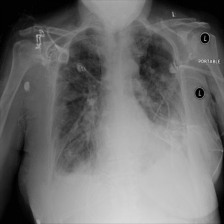}    & 
\includegraphics[width=0.1\linewidth]{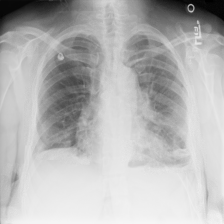}    & 
\includegraphics[width=0.1\linewidth]{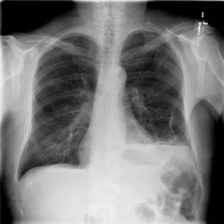}    &                          
\includegraphics[width=0.1\linewidth]{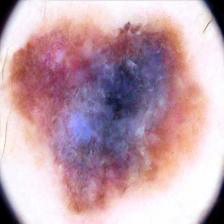}    & 
\includegraphics[width=0.1\linewidth]{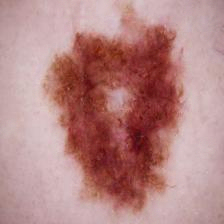}    & 
\includegraphics[width=0.1\linewidth]{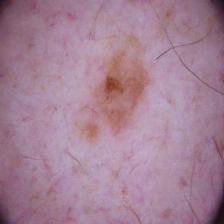}                     
\\
\raisebox{2pt}{ \scalebox{.5}{\begin{sideways}{\FastDiMETwoPlus}\end{sideways}} }&
\includegraphics[width=0.1\linewidth]{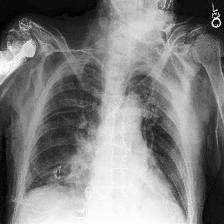}  & 
\includegraphics[width=0.1\linewidth]{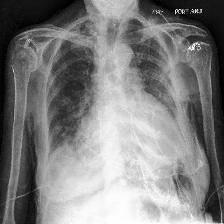}  & 
\includegraphics[width=0.1\linewidth]{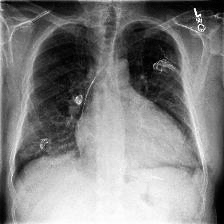}  &                           
\includegraphics[width=0.1\linewidth]{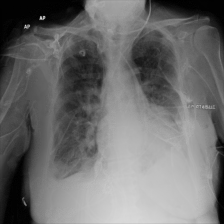}  & 
\includegraphics[width=0.1\linewidth]{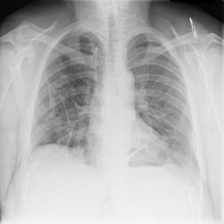}  & 
\includegraphics[width=0.1\linewidth]{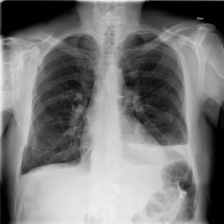}  &                          
\includegraphics[width=0.1\linewidth]{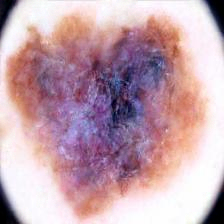}  & 
\includegraphics[width=0.1\linewidth]{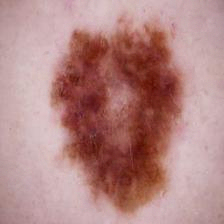}  & 
\includegraphics[width=0.1\linewidth]{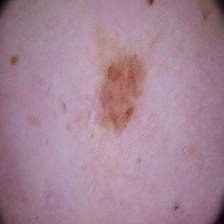} 
\end{tabular}
   \caption{\textbf{Shortcut counterfactuals for medical datasets.} The shortcuts are highlighted with orange circles or boxes in the original images.}
   \label{fig:medical_case_samples}
\end{figure}

\begin{figure}[t]
\setlength{\tabcolsep}{2pt}
\centering
\scalebox{0.5}{
\begin{tabular}{ccccc|ccccc}
    \footnotesize Original & \footnotesize ACE $\ell_1$  & \footnotesize DiME & \footnotesize \FastDiME & \footnotesize \FastDiMETwoPlus 
  &
    \footnotesize Original & \footnotesize ACE $\ell_1$  & \footnotesize DiME & \footnotesize \FastDiME & \footnotesize \FastDiMETwoPlus\\

  \includegraphics[width=0.18\linewidth]{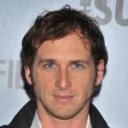} & \includegraphics[width=0.18\linewidth]{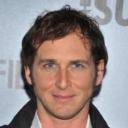} &
   \includegraphics[width=0.18\linewidth]{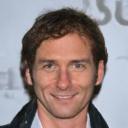} &
   \includegraphics[width=0.18\linewidth]{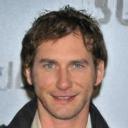} &
   \includegraphics[width=0.18\linewidth]{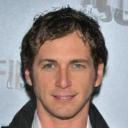}
   &
   \includegraphics[width=0.18\linewidth]{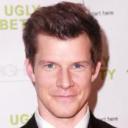} & \includegraphics[width=0.18\linewidth]{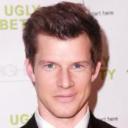}
   &\includegraphics[width=0.18\linewidth]{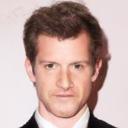}
   &\includegraphics[width=0.18\linewidth]{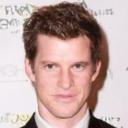}
   &\includegraphics[width=0.18\linewidth]{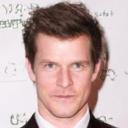}
   \\
   
   \includegraphics[width=0.18\linewidth]{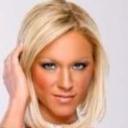} & \includegraphics[width=0.18\linewidth]{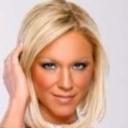}
   &\includegraphics[width=0.18\linewidth]{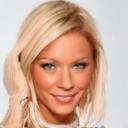}
   &\includegraphics[width=0.18\linewidth]{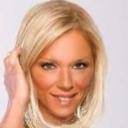}
   &\includegraphics[width=0.18\linewidth]{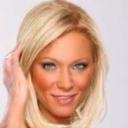}
   &
   \includegraphics[width=0.18\linewidth]{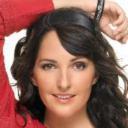} & \includegraphics[width=0.18\linewidth]{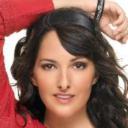}
    &\includegraphics[width=0.18\linewidth]{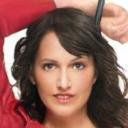}
    &\includegraphics[width=0.18\linewidth]{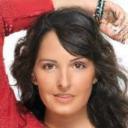}
    &\includegraphics[width=0.18\linewidth]{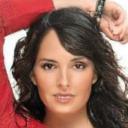}
   \\
    \includegraphics[width=0.18\linewidth]{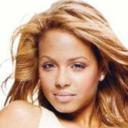} & \includegraphics[width=0.18\linewidth]{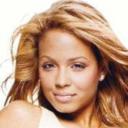}
   &\includegraphics[width=0.18\linewidth]{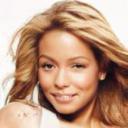}
   &\includegraphics[width=0.18\linewidth]{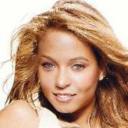}
   &\includegraphics[width=0.18\linewidth]{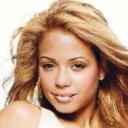}
   &
    \includegraphics[width=0.18\linewidth]{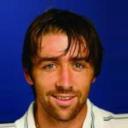} & \includegraphics[width=0.18\linewidth]{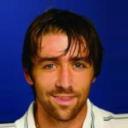}
    &\includegraphics[width=0.18\linewidth]{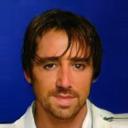}
    &\includegraphics[width=0.18\linewidth]{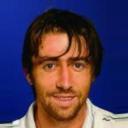}
    &\includegraphics[width=0.18\linewidth]{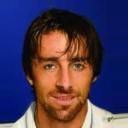}
\\
\multicolumn{5}{c}{ \scalebox{1.7}{No smile $\rightarrow$ Smile.} } &
\multicolumn{5}{c}{ \scalebox{1.7}{Smile $\rightarrow$ No smile.} } \\
\end{tabular}
}
   \caption{\textbf{CelebA counterfactual explanations for the `smile' attribute.} }
   \label{fig:celeba}
\end{figure}

\paragraph{Results.}
\cref{tab:celeba-main} and \cref{tab:medical_main} list the results of the counterfactual explanation experiments on CelebA and medical datasets, respectively. For most of the datasets, \FastDiME, and its variants outperform DiME in most of the metrics, while it remains competitive against ACE in terms of image quality. ACE achieves a significantly better COUT than our method as it is sufficient to cross the decision boundary, whereas we explicitly maximize the counterfactual class probability. This can be observed in the BKL, MAD, and FR metrics that demonstrate that our method can indeed produce samples that maximize the counterfactual class probability. Further illustrating this point,~\cref{fig:celeba} shows exemplary results on CelebA, demonstrating how our methods tend to produce stronger counterfactual `smiling' impressions compared to ACE.
\cref{fig:medical_case_samples} shows generated shortcut counterfactuals on the medical datasets where the shortcut feature has been successfully removed from the images without significant changes to the remainder of the image.

\begin{table}[h]
    \centering
    \caption{\textbf{Complexity, time, and memory consumption.} \FastDiMEnonMask{} refers to our method without self-optimized mask scheme. $T$ denotes the number of diffusion steps and $N$ is the number of adversarial attack update steps in ACE. Total Time (hours) is for a random subset of 1000 CelebA images, Batch Time (seconds) is calculated with a batch of 5, and GPU MU is in MiB.}
    \label{tab:celeba-main2}
    \footnotesize
    
\scalebox{0.7}{\begin{tabular}{c|cccc} \toprule
        Method        & Complexity & Batch Time & Total Time & GPU MU \\ \midrule
        DiME          & $\mathcal{O}(T^2)$& 218.8\tiny{$\pm$72.6} 
        & 12:10:08 & \textbf{1.6K} 
        \\
         ACE $\ell_1$  & $\mathcal{O}(NT)$& 41.8\tiny{$\pm$0.7} 
         & 02:17:46 & 13.1K 
         \\
        ACE $\ell_2$  & $\mathcal{O}(NT)$& 42.9\tiny{$\pm$0.1} 
        & 02:18:43 &
        13.1K  \\ 
        
        GMD & $\mathcal{O}(T)$& 31.4\tiny{$\pm$6.6} & 01:45:45 & 4.2K  \\ \midrule

        \FastDiMEnonMask & $\mathcal{O}(T)$& \textbf{10.6\tiny{$\pm$4.0}} & \textbf{00:35:56} &\textbf{1.6K} \\
        \FastDiME & $\mathcal{O}(T)$ & \textit{13.6\tiny{$\pm$5.1}} & \textit{00:45:55} &\textbf{1.6K} \\
        
        \FastDiMETwo & $\mathcal{O}(T)$ & 23.1\tiny{$\pm$9.3}  & 01:17:40 &\textbf{1.6K} \\
        \FastDiMETwoPlus & $\mathcal{O}(T)$& 27.0\tiny{$\pm$10.4} & 01:30:38 &\textbf{1.6K} 
         
         \\\bottomrule 
    \end{tabular}
    }
    
\end{table}

\subsection{Efficiency analysis}
\label{sec:efficiencyanalysis}
We compare diffusion-based methods in terms of inference time and GPU memory usage (MU) in MiB. Using a random subset of 1000 images from CelebA, we generated counterfactuals for the `smile' attribute with a batch size of 5 on an RTX 5000 Turing 16 GB NVIDIA GPU. We report the average batch time in seconds and the total time in hours together with the theoretical complexity in~\cref{tab:celeba-main2}. Note that we could not fit batch sizes larger than 10 for ACE~\cite{jeanneret2023adversarial}.

\subsection{Shortcut learning detection pipeline}
We train ResNet-18~\cite{he2016deep} classifiers $f_k$ initialised with ImageNet~\cite{deng2009imagenet} pre-trained weights on each of the datasets $\mathcal{D}_{k} , k=\{100,75,50\}$ and evaluate them on the three curated test sets, $\mathrm{test}_{k}$ $\mathrm{test}_u$, and  $\mathrm{test}^c_u$. The task labels and suspected shortcuts are shown in~\cref{tab:shortcuts_res}.

\paragraph{Evaluation criteria.}
To \emph{quantify} the shortcut learning effect using counterfactuals, we measure the differences in model predictions resulting from counterfactually changing $s$ of the test images. To this end, we measure MAD between original images ($\mathrm{test}_{u}$) and shortcut counterfactuals ($\mathrm{test}^c_u$), and their Mean confidence Difference (MD) across two subtests $X_{s=0}$ and $X_{s=1}$ according to their true shortcut label $s$.
Furthermore, in order to \emph{validate} the extent of shortcut learning apart from the unbalanced dataset setting, 
we also evaluate each of the classifiers $f_k$ trained on different correlation levels $k$ with the shortcut feature $s$ using Area Under the Receiver Operating Characteristic (AUROC). 

\paragraph{Results.}
\cref{tab:shortcuts_res} and \cref{fig:shortcuts_res} shows the results of our shortcut learning detection experiments.
The magnitude of model prediction changes in terms of MAD and MD metrics between original test images and \FastDiME{} shortcut counterfactuals correctly signifies stronger reliance of the models on the shortcut feature for the models trained on the more strongly biased training sets. 
The significant difference between $test_k$ and $test_u$ in terms of AUROC also indicates that the trained classifiers indeed learn to exploit the shortcut, with the severity of shortcut learning being correlated with the strength $k\%$ of the association in the training set. 
\cref{fig:shortcuts_res}, demonstrates that the proposed pipeline is indeed appropriate for detecting and quantifying shortcut learning in practice.
Notice that AUROC is only used as an indicator, as it is not a reliable measurement for shortcut learning. The difference between the AUROC of $test_k$ and $test_u$ can only tell whether the shortcut is correlated with the target label, but it can not exclude other potential reasons and cannot reveal a causal relationship. If pacemakers are highly correlated with text markers in X-rays, the AUROC results cannot determine which is the shortcut.
It should also be noted that the proposed shortcut detection pipeline is designed to obviate the need for diagnostic labels for new data within $test_u$, thereby facilitating any new databases after the trained counterfactual pipeline.

\begin{table*}[t]
    \centering
    \scalebox{0.56}{
    \centering
    \begin{tabular}[b]{c | cc | c | c c c | c c c}
    \toprule
    \multirow{ 2}{*}{Dataset} & \multirow{ 2}{*}{Task Label} & \multirow{ 2}{*}{Shortcut} & \multirow{ 2}{*}{Train Set} & \multicolumn{3}{c|}{AUROC} &\multicolumn{3}{c}{Shortcut Detection Metrics} \\
    &&&&$\mathrm{test}_k$ & $\mathrm{test}_u$ & $\mathrm{test}^c_u$ & MAD & MD($_{s=1}$) & MD($_{s=0}$) \\ 
    \midrule

    \multirow{3}{*}{{\footnotesize CheXpert}} & \multirow{3}{*}{{ cardiomegaly}} & \multirow{3}{*}{{pacemaker}}
    
    & $\mathcal{D}_{100}$  & 0.98 & 0.58 & 0.63& 0.36 &0.40 &-0.26\\ 

    &&& $\mathcal{D}_{75}$  & 0.80 & 0.71 & 0.74&0.15 &0.14&-0.03 \\ 
    &&& $\mathcal{D}_{50}$  & 0.72 & 0.73 &0.74 & 0.12& -0.01 & -0.02\\
     \midrule 
    \multirow{3}{*}{{\footnotesize NIH}} &\multirow{3}{*}{{pneumothorax}} & \multirow{3}{*}{{chest drain}}
    
    & $\mathcal{D}_{100}$  & 0.98 & 0.65 &0.65 & 0.13 & 0.04& -0.22\\
    &&& $\mathcal{D}_{75}$  &0.87 &0.73&0.71 & 0.07 & 0.01 &-0.10 \\
    &&& $\mathcal{D}_{50}$   &0.70&0.71& 0.70 & 0.03 & -0.01 &  -0.02\\
   \midrule 
    \multirow{3}{*}{{\footnotesize ISIC}} & \multirow{3}{*}{{ malignant}} & \multirow{3}{*}{{ruler markers}}
    
    & $\mathcal{D}_{100}$  & 0.98 & 0.70 & 0.79 & 0.19& 0.15 & -0.20\\
    &&& $\mathcal{D}_{75}$  & 0.86 & 0.82 & 0.82 & 0.10  & 0.01 & -0.09\\
    &&& $\mathcal{D}_{50}$ & 0.87 & 0.85 & 0.81 & 0.09 & -0.04 & -0.03\\
    \midrule 

    \multirow{3}{*}{{\footnotesize CelebA}} & \multirow{3}{*}{{ age}} & \multirow{3}{*}{{smile}}

    & $\mathcal{D}_{100}$  & 0.98  & 0.65 & 0.73 & 0.32  & 0.36 & -0.27 \\
    &&& $\mathcal{D}_{75}$  &  0.91 & 0.84 & 0.83 & 0.17&  0.18 &  -0.04 \\
    &&& $\mathcal{D}_{50}$  & 0.87 & 0.87 &  0.85 & 0.14 & 0.09 &  0.04\\

    \bottomrule
    \end{tabular}
    }
    \qquad
    \scalebox{0.75}{
     \includegraphics[width=0.39\linewidth]{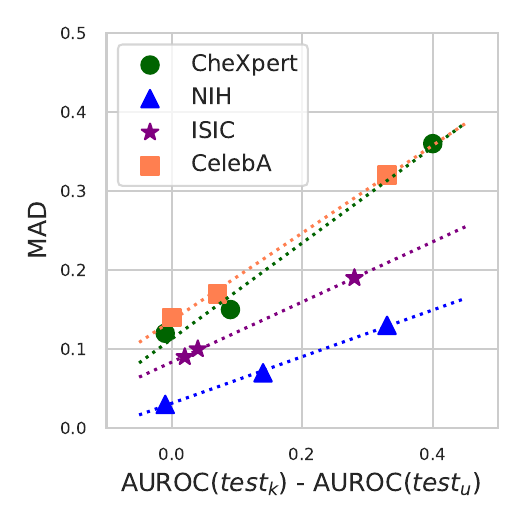}
    }
    \captionlistentry[figure]{}
    \label{fig:shortcuts_res}
    \captionsetup{labelformat=andfigure}
    \caption{\textbf{Results of proposed shortcut detection pipeline.} The table (\textit{left}) details the prediction performance on the distinct test sets, $test_k$ (same distribution as trainset) and $test_u$ (unseen balanced set), alongside the associated shortcut detection metrics. The observed performance discrepancy between $test_k$ and $test_u$ quantitatively indicates the extent to which shortcuts affect the main task's predictive accuracy, which correlates with the shortcut detection metrics, as illustrated in the figure (\textit{right}). 
    Although AUROC indicates the presence of shortcuts, it is not a reliable measurement, as it only shows the correlation but no causality.
    }
    \label{tab:shortcuts_res}
\end{table*}

\section{Discussion, limitations and conclusion}

\paragraph{Adversarially vulnerable classifier.}

\begin{figure}[t]
\setlength{\tabcolsep}{2pt}
\centering
\scalebox{0.5}{
\begin{tabular}{ccccc|ccccc}
  \centering
   \footnotesize Original & \footnotesize ACE $\ell_1$  & \footnotesize DiME & \footnotesize \FastDiME & \footnotesize \FastDiMETwoPlus
   &
   \footnotesize Original & \footnotesize ACE $\ell_1$  & \footnotesize DiME & \footnotesize \FastDiME & \footnotesize \FastDiMETwoPlus\\

     \includegraphics[width=0.18\linewidth]{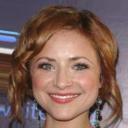}
&\includegraphics[width=0.18\linewidth]{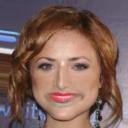}
   &\includegraphics[width=0.18\linewidth]{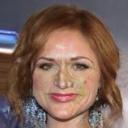}
   &\includegraphics[width=0.18\linewidth]{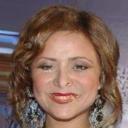}
    &\includegraphics[width=0.18\linewidth]{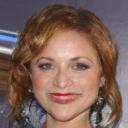}&
        \includegraphics[width=0.18\linewidth]{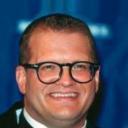}
   &\includegraphics[width=0.18\linewidth]{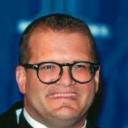}
   &\includegraphics[width=0.18\linewidth]{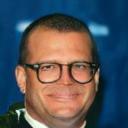}
   &\includegraphics[width=0.18\linewidth]{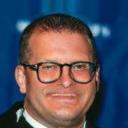}
    &\includegraphics[width=0.18\linewidth]{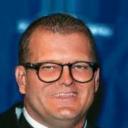} \\

   \includegraphics[width=0.18\linewidth]{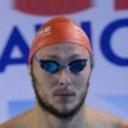}
   &\includegraphics[width=0.18\linewidth]{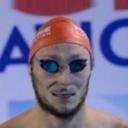}
   &\includegraphics[width=0.18\linewidth]{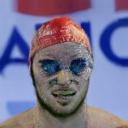}
   &\includegraphics[width=0.18\linewidth]{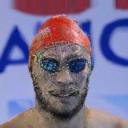}
    &\includegraphics[width=0.18\linewidth]{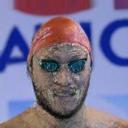}
    &
    \includegraphics[width=0.18\linewidth]{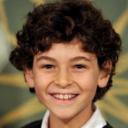}
   &\includegraphics[width=0.18\linewidth]{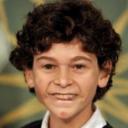}
   &\includegraphics[width=0.18\linewidth]{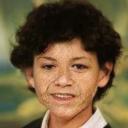}
   &\includegraphics[width=0.18\linewidth]{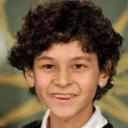}
    &\includegraphics[width=0.18\linewidth]{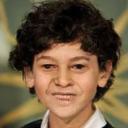}

\end{tabular}
}

   \caption{\textbf{Limitations for valid counterfactuals.}
   The left-top case shows a hard example (smile $\rightarrow$ no smile) where all method fails to make semantically meaningful changes. In the left-bottom (no smile $\rightarrow$ smile), all DiME-based methods fail to produce a suitable sample, while ACE's result appears natural but lacks a noticeable smile alteration. In the right-top (smile $\rightarrow$ no smile), we observe similar performance for ACE, while our 2-step approach offers a visually better result. Right-bottom case (smile $\rightarrow$ no smile) shows an example where all methods change more than the targeted smile.
}
   \label{fig:celeba_limitation}
\end{figure}

We find, across methods, that the generalization performance of the classifier used for generating counterfactuals is crucial. In addition to classifying samples within the data distribution, the counterfactual generation process introduces out-of-distribution generated samples. This causes most counterfactual failures, as the model ceases to guide once the prediction flips. Examples are demonstrated in \cref{fig:celeba_limitation}. A potential solution would be to use an adversarially robust classifier, to obtain more informative gradients~\cite{Etmann2019} and realistic generated images~\cite{Santurkar2019, Boreiko2022, Neuhaus2023}. This solution mostly applies for counterfactual image synthesis, \emph{not} for counterfactual explanations, as for the latter the goal is to explain any classifier, irrespective of its robustness.

\paragraph{Denoised images and masking parameter $\tau_w$.} 

As expected, our efficient gradient estimation produces denoised images $\bar{x}^c_t$ that tend to be more blurred during the early time steps compared to the expensive generated images $\hat{x}^c_t$ from DiME. Yet, as denoised images diminish noise levels and shortcut features are often highly localized and less sensitive to blurriness, our counterfactuals do not degrade image quality. To further improve them, we apply our self-optimized mask scheme as long as $\bar{x}^c_t$ is not blurred, which is around $\frac{\tau}{2}$. We explored quantitatively and quantitatively the effects of our self-optimized masking in the Appendix. 

\paragraph{Unbiased shortcut generation.}
One important limitation of our work is that it depends on unbiased shortcut classification and image generation -- image generation models have also been shown to reproduce (potentially spurious) correlations from training~\cite{Luccioni2023}.
A counterfactual generator might thus not only change the shortcut feature but also discriminative features, e.g., the disease label. We mitigate this by training our generative shortcut classifier for medical datasets only on diseased samples, and by using self-optimized masks to limit the spatial extent of the counterfactual changes.
Our method can sometimes result in counterfactuals that change more than the targeted shortcut (see \cref{fig:celeba_limitation}).
Further development of the self-optimized masking and reducing the effect of spurious correlations on the generative shortcut model remain important avenues for future work.
Note that it is often easier to train an unbiased shortcut classifier as opposed to for: Shortcut features are typically easier to predict, which is precisely what makes them vulnerable.

\paragraph{Using shortcut counterfactuals to mitigate shortcut learning.}
While we focus on generating the shortcut counterfactuals and using them to \emph{detect} shortcut learning, they also have the potential to \emph{mitigate} shortcut learning by augmenting the training set with samples that have the shortcut feature added or removed. 
Similar approaches have been successful in other domains~\cite{pakzad2022circle, Chang2021, Balashankar2021, Nauta2021, Qiang2022}.
\paragraph{Conclusion.}

We present a general and fast method for diffusion-based counterfactual explanations, which is approximately up to 20 times faster than the existing methods while preserving comparable counterfactual quality. We further demonstrate how our method can be used for generating counterfactuals with and without suspected shortcuts in both medical datasets and CelebA with convincing visual results. Based on this, we introduce a novel pipeline to detect shortcut learning in practice.

\newpage
\section*{Acknowledgements}
Work on this project was partially funded by the Independent Research Fund Denmark (DFF, grant number 9131-00097B), the Pioneer Centre for AI (DNRF grant nr P1), the DIREC project EXPLAIN-ME (9142-00001B), and the Novo Nordisk Foundation through the Center for Basic Machine Learning Research in Life Science (MLLS, grant NNF20OC0062606). The funding agencies had no influence on the writing of the manuscript nor on the decision to submit it for publication.

% ---- Bibliography ----
%
% BibTeX users should specify bibliography style 'splncs04'.
% References will then be sorted and formatted in the correct style.
%
\bibliographystyle{splncs04}
\bibliography{main}

\newpage
\appendix

{\Large{\textbf{Appendix}}}

\section{Implementation details}
\label{details}

\subsection{Medical datasets}
\paragraph{Shortcut annotations.} The annotations of chest drains from NIH are crow-sourced from both radiologists and non-experts~\cite{damgaard2023augmenting,oakden2020hidden}, and pacemakers are labeled using LabelMe~\cite{Wada_Labelme_Image_Polygonal}. Rulers are also annotated using LabelMe~\cite{Wada_Labelme_Image_Polygonal}. Our shortcut annotations for CheXpert-pacemaker and ISIC-Ruler marker are public at \href{https://github.com/nina-weng/FastDiME_Med}{github.com/nina-weng/FastDiME\_Med}.

\paragraph{Dataset size.}
Notably, the test sets for the medical cases in evaluation of counterfactual explanations are relatively small (CheXpert-PM: 202; NIH-Drain: 164; ISIC-Ruler: 173), containing exclusively cases with existing shortcut features. This selection is more meaningful for medical applications, where removing (\textit{not adding}) potential shortcut features is often more useful.

\subsection{Counterfactual loss and FastDiME hyperparameters}

\paragraph{Counterfactual loss computation.} We leverage the denoised image to compute the counterfactual loss.
For all medical datasets, $L_1$ loss is computed between the denoised image $\bar{x}^c_{t}$ and the original image $x_0$, while we do not include a perceptual loss. For CelebA, $L_{perc}$ is computed between the denoised image $\bar{x}^c_{t}$ and the original image $x_0$, and implemented with a pre-trained VGG-19~\cite{simonyan2014very} as feature extractor, while $L_1$ loss is computed between the noisy image~$x^c_t$ and the original image $x_0$, following the original implementation of DiME~\cite{jeanneret2022diffusion}.

\paragraph{Hyperparameters.}
For CelebA, we use the same hyperameters as in DiME~\cite{jeanneret2022diffusion}, i.e., $\tau = 60$ out of 200 re-spaced time-steps, $\lambda_{p}=30$ and $\lambda_{1} = 0.05$ and $\lambda_{c} \in \{8,10,15\}$. For our self-optimized masking scheme, we normalize the masks, use a threshold of 0.15, and perform a dilation operation to set the mask as a square with a width and height of 5 pixels, similar to ACE~\cite{jeanneret2023adversarial}. For all medical datasets, we set $\tau = 160$ out of 400 re-spaced time-steps, and $\lambda_{c} = 1 $ and $\lambda_{1} = 50 $ and $\lambda_{p} = 0$, and use a threshold of 0.15 and a dilation of 21 for our masks. For all datasets, we set $\tau_w = \frac{\tau}{2}$.

\subsection{Experiments on shortcut detection} 

In this section, we present the proposed shortcut evaluation metrics as well as details and design choices for the shortcut detection experiments.

\paragraph{Shortcuts detection metrics.}

We define the Mean Absolute confidence Difference (MAD) across all samples as follows, 
\begin{equation}
    \mathrm{MAD} = \frac{1}{N} \Sigma_{i=1}^N |f(x_i) - f(x_i^c)|
\end{equation} 
where $x_i$ represents the original image (contained in $\mathrm{test}_u$) and $x_i^c$ its shortcut counterfactual (contained in $\mathrm{test}_c$). 
MAD is within the range of $[0,1]$, where $1$ indicates the complete flip in confidence and $0$ suggests no changes at all in inference. 

Similarly, the Mean confidence Difference (MD)
is defined as follows,

\begin{equation}
    \mathrm{MD} = \frac{1}{N} \Sigma_{i=1}^N f(x_i) - f(x_i^c)
\end{equation} 
We compute MD across the two subsets $X_{s=0}$ and $X_{s=1}$ of the samples in $\mathrm{test}_u$ according to their true shortcut labels. MD is within the range of $[-1,1]$, where $1$ represents the drop in confidence level, $-1$ represents the contrary, and $0$ indicates no changes.

\paragraph{Experimental design for shortcut detection.}
Table~\ref{tab:shortcuts_detection_settings} lists the task labels, shortcuts, and sizes for all datasets used in this work. It should be noted that, for the NIH dataset, since we do not have shortcut annotations for non-disease samples, we first trained a DenseNet~\cite{huang2017densely} model using 3543 annotated samples, and we then inferred the confidence level for all samples. By thresholding with $0.8$ and $0.01$, we selected the samples with high/low confidence, which are regarded as more trustworthy predictions, together with the annotated ones. When selecting samples for dataset preparation, we prioritize the samples with higher confidence levels. The predicted chest drain labels used here will be publicly available.

\begin{table}[t]
    \centering
    \caption{\textbf{Experimental design for shortcut detection.} Dataset size, main task, and shortcuts design for the shortcuts detection experiments.}
    \label{tab:shortcuts_detection_settings}

    \footnotesize
    \scalebox{0.89}{
    \begin{tabular}{c | c c  | c c c c}
    \toprule
    \multirow{ 2}{*}{Data} & \multirow{ 2}{*}{Task Label} & \multirow{ 2}{*}{Shortcut} &\multicolumn{3}{c}{Dataset Size} \\
    &&& $train$ & $test_k$ & $test_{u/c}$ \\ 
    \midrule

    CheXpert & cardiomegaly & pacemaker & 1395 & 349 & 388 \\
    NIH & pneumothorax & chest drain$^\dag$  & 2771 &693 & 1728\\
    ISIC & malignant$^*$ & ruler markers & 814 & 204 & 508\\
    CelebA & age & smile & 10000 & 1000 & 1000\\

    \bottomrule
    
    \end{tabular}
    }
    
    \begin{tablenotes}
        \footnotesize
        \item $^\dag$ Chest drain labels contains both non-expert annotations~\cite{damgaard2023augmenting} and  estimated labels by DenseNet.\\
        \item $^*$ Malignant lesions include melanoma, basal cell carcinoma, actinic keratosis/bowen's disease/keratoacanthoma/squamous cell carcinoma, dermatofibromas, and vascular lesions. For more information about the diseases, please refer to~\cite{codella2019skin}.
    \end{tablenotes}
    
\end{table}

\paragraph{Training details.} 
We train the ResNet-18~\cite{he2016deep} classifiers initialized with Image-Net~\cite{deng2009imagenet} pre-trained weights. For medical datasets, the classifiers are trained for a maximum of 50 epochs with a learning rate of $1\times10^{-6}$, a weight decay of 0.05, an early-stop mechanism monitoring validation loss with patience of 10 epochs, and Adam~\cite{kingma2014adam} optimizer. For CelebA, we train classifiers for 20 epochs using a learning rate of $1\times10^{-5}$, a weight decay of 0.05, and Adam~\cite{kingma2014adam} optimizer.

\section{Qualitative results}
\label{QualitativeResults}
\begin{figure*}
\centering
\scalebox{0.7}{
\begin{tabular}{c|ccccccc}
   & Original & ACE $\ell_1$& DiME & \FastDiME & \FastDiMETwo & \FastDiMETwoPlus \\

   & \includegraphics[scale=0.6]{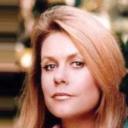}  & \includegraphics[scale=0.6]{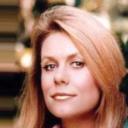} & \includegraphics[scale=0.6]{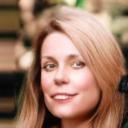} & \includegraphics[scale=0.6]{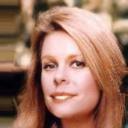} & \includegraphics[scale=0.6]{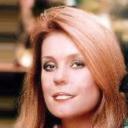} & \includegraphics[scale=0.6]{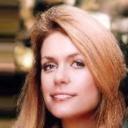} \\
   
   \multirow{2}{*}{\rotatebox{90}{no smile $\rightarrow$ smile}}  & \includegraphics[scale=0.6]{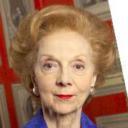}  & \includegraphics[scale=0.6]{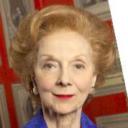} & \includegraphics[scale=0.6]{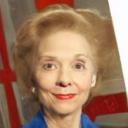} & \includegraphics[scale=0.6]{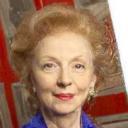} & \includegraphics[scale=0.6]{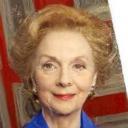} & \includegraphics[scale=0.6]{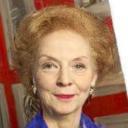} \\

   & \includegraphics[scale=0.6]{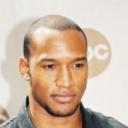}  & \includegraphics[scale=0.6]{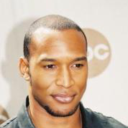} & \includegraphics[scale=0.6]{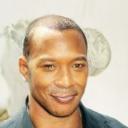} & \includegraphics[scale=0.6]{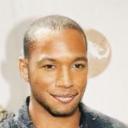} & \includegraphics[scale=0.6]{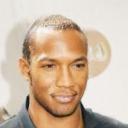} & \includegraphics[scale=0.6]{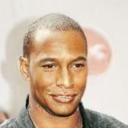} \\

   & \includegraphics[scale=0.6]{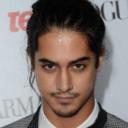}  & \includegraphics[scale=0.6]{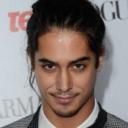} & \includegraphics[scale=0.6]{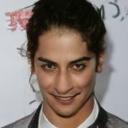} & \includegraphics[scale=0.6]{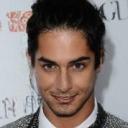} & \includegraphics[scale=0.6]{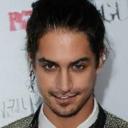} & \includegraphics[scale=0.6]{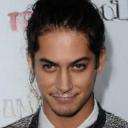} \\

   & \includegraphics[scale=0.6]{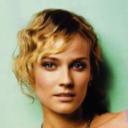}  & \includegraphics[scale=0.6]{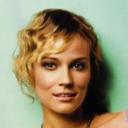} & \includegraphics[scale=0.6]{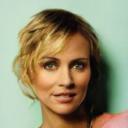} & \includegraphics[scale=0.6]{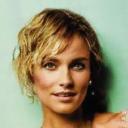} & \includegraphics[scale=0.6]{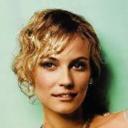} & \includegraphics[scale=0.6]{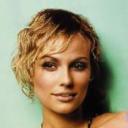} \\ \cmidrule{1-7}
   
    & \includegraphics[scale=0.6]{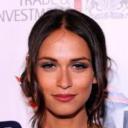}  & \includegraphics[scale=0.6]{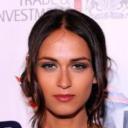} & \includegraphics[scale=0.6]{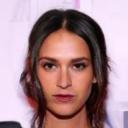} & \includegraphics[scale=0.6]{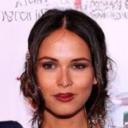} & \includegraphics[scale=0.6]{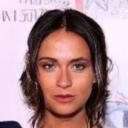} & \includegraphics[scale=0.6]{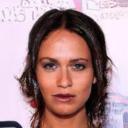} \\

  \multirow{2}{*}{\rotatebox{90}{smile $\rightarrow$ no smile}} & \includegraphics[scale=0.6]{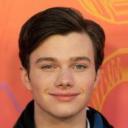}  & \includegraphics[scale=0.6]{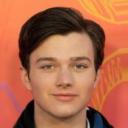} & \includegraphics[scale=0.6]{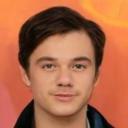} & \includegraphics[scale=0.6]{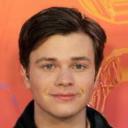} & \includegraphics[scale=0.6]{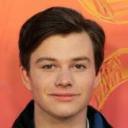} & \includegraphics[scale=0.6]{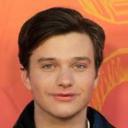} \\

   & \includegraphics[scale=0.6]{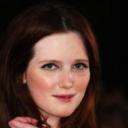}  & \includegraphics[scale=0.6]{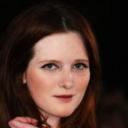} & \includegraphics[scale=0.6]{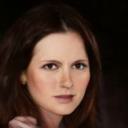} & \includegraphics[scale=0.6]{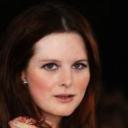} & \includegraphics[scale=0.6]{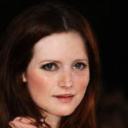} & \includegraphics[scale=0.6]{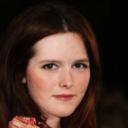} \\

   & \includegraphics[scale=0.6]{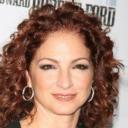}  & \includegraphics[scale=0.6]{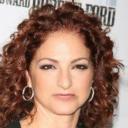} & \includegraphics[scale=0.6]{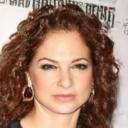} & \includegraphics[scale=0.6]{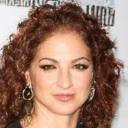} & \includegraphics[scale=0.6]{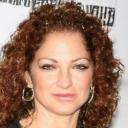} & \includegraphics[scale=0.6]{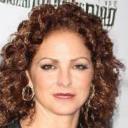} \\ 

   & \includegraphics[scale=0.6]{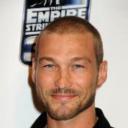}  & \includegraphics[scale=0.6]{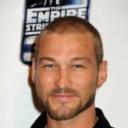} & \includegraphics[scale=0.6]{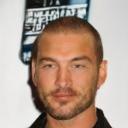} & \includegraphics[scale=0.6]{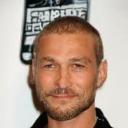} & \includegraphics[scale=0.6]{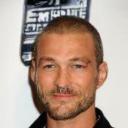} & \includegraphics[scale=0.6]{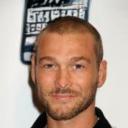} \\

\end{tabular}
}
\caption{\textbf{CelebA}. Comparisons for the `smile' attribute.}
\label{fig:celebasmile}
\end{figure*}

\begin{figure*}
\centering
\scalebox{0.7}{
\begin{tabular}{c|ccccccc}
   & Original & ACE $\ell_1$& DiME & \FastDiME & \FastDiMETwo & \FastDiMETwoPlus \\

   & \includegraphics[scale=0.6]{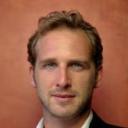}  & \includegraphics[scale=0.6]{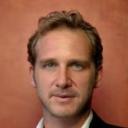} & \includegraphics[scale=0.6]{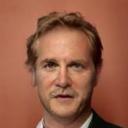} & \includegraphics[scale=0.6]{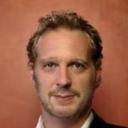} & \includegraphics[scale=0.6]{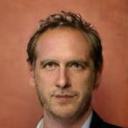} & \includegraphics[scale=0.6]{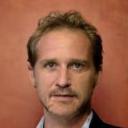} \\
   
   \multirow{2}{*}{\rotatebox{90}{young $\rightarrow$ old}}  & \includegraphics[scale=0.6]{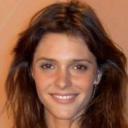}  & \includegraphics[scale=0.6]{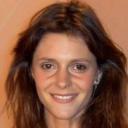} & \includegraphics[scale=0.6]{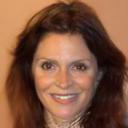} & \includegraphics[scale=0.6]{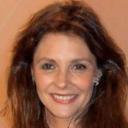} & \includegraphics[scale=0.6]{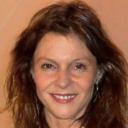} & \includegraphics[scale=0.6]{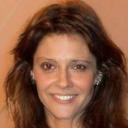} \\

    & \includegraphics[scale=0.6]{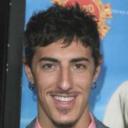}  & \includegraphics[scale=0.6]{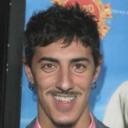} & \includegraphics[scale=0.6]{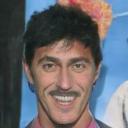} & \includegraphics[scale=0.6]{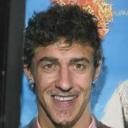} & \includegraphics[scale=0.6]{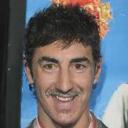} & \includegraphics[scale=0.6]{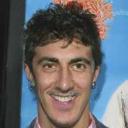} \\

  & \includegraphics[scale=0.6]{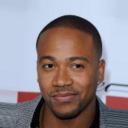}  & \includegraphics[scale=0.6]{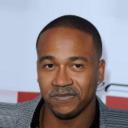} & \includegraphics[scale=0.6]{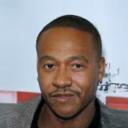} & \includegraphics[scale=0.6]{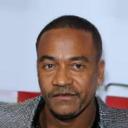} & \includegraphics[scale=0.6]{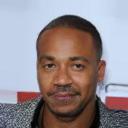} & \includegraphics[scale=0.6]{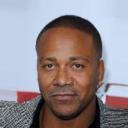} \\

    & \includegraphics[scale=0.6]{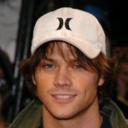}  & \includegraphics[scale=0.6]{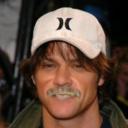} & \includegraphics[scale=0.6]{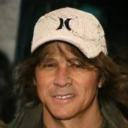} & \includegraphics[scale=0.6]{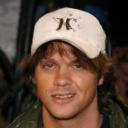} & \includegraphics[scale=0.6]{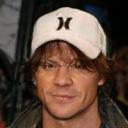} & \includegraphics[scale=0.6]{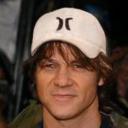} \\ \cmidrule{1-7}

    & \includegraphics[scale=0.6]{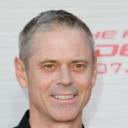}  & \includegraphics[scale=0.6]{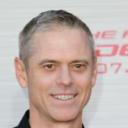} & \includegraphics[scale=0.6]{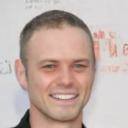} & \includegraphics[scale=0.6]{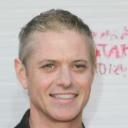} & \includegraphics[scale=0.6]{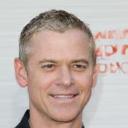} & \includegraphics[scale=0.6]{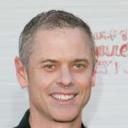} \\

    \multirow{2}{*}{\rotatebox{90}{old $\rightarrow$ young}} & \includegraphics[scale=0.6]{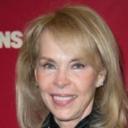}  & \includegraphics[scale=0.6]{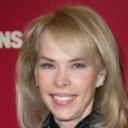} & \includegraphics[scale=0.6]{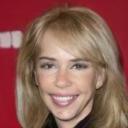} & \includegraphics[scale=0.6]{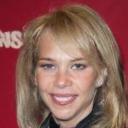} & \includegraphics[scale=0.6]{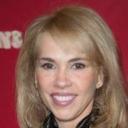} & \includegraphics[scale=0.6]{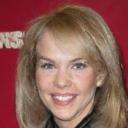} \\

    & \includegraphics[scale=0.6]{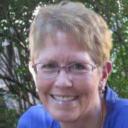}  & \includegraphics[scale=0.6]{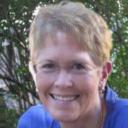} & \includegraphics[scale=0.6]{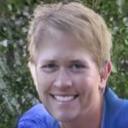} & \includegraphics[scale=0.6]{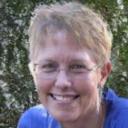} & \includegraphics[scale=0.6]{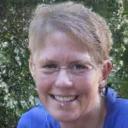} & \includegraphics[scale=0.6]{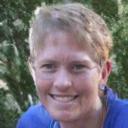} \\

    & \includegraphics[scale=0.6]{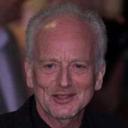}  & \includegraphics[scale=0.6]{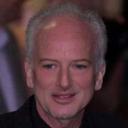} & \includegraphics[scale=0.6]{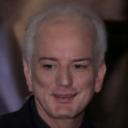} & \includegraphics[scale=0.6]{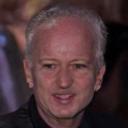} & \includegraphics[scale=0.6]{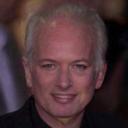} & \includegraphics[scale=0.6]{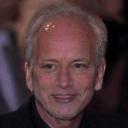} \\

    & \includegraphics[scale=0.6]{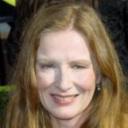}  & \includegraphics[scale=0.6]{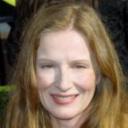} & \includegraphics[scale=0.6]{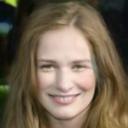} & \includegraphics[scale=0.6]{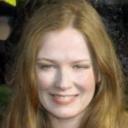} & \includegraphics[scale=0.6]{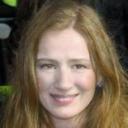} & \includegraphics[scale=0.6]{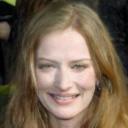} \\

\end{tabular}
}
\caption{\textbf{CelebA}. Comparisons for the `age' attribute.}
\label{fig:celebaage}
\end{figure*}

\begin{figure*}
\centering
\scalebox{0.65}{
\begin{tabular}{ccccccc}
   & Original & DiME & \FastDiMEnonMask & \FastDiME & \FastDiMETwo & \FastDiMETwoPlus \\

   & \includegraphics[scale=0.4]{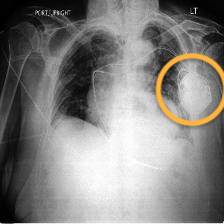}  & \includegraphics[scale=0.4]{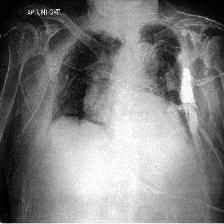} & \includegraphics[scale=0.4]{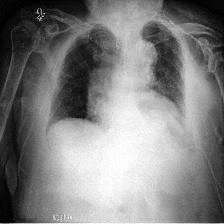} & \includegraphics[scale=0.4]{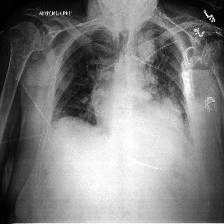} & \includegraphics[scale=0.4]{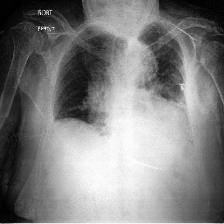} & \includegraphics[scale=0.4]{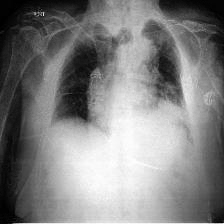} \\

   & \includegraphics[scale=0.4]{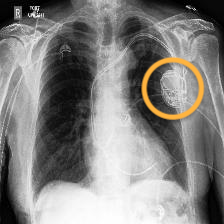}  & \includegraphics[scale=0.4]{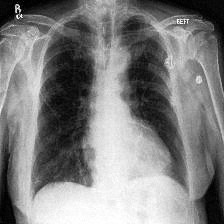} & \includegraphics[scale=0.4]{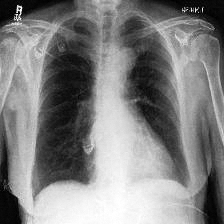} & \includegraphics[scale=0.4]{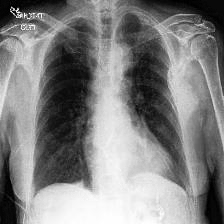} & \includegraphics[scale=0.4]{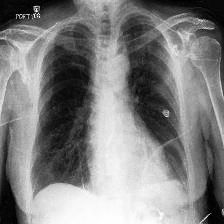} & \includegraphics[scale=0.4]{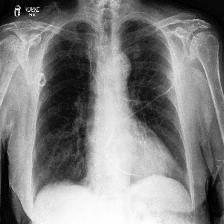} \\
   
    & \includegraphics[scale=0.4]{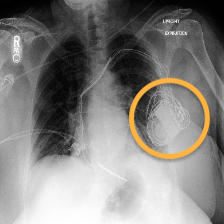}  & \includegraphics[scale=0.4]{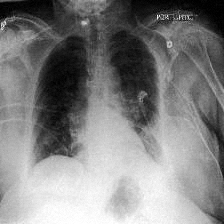} & \includegraphics[scale=0.4]{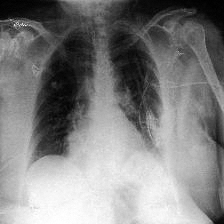} & \includegraphics[scale=0.4]{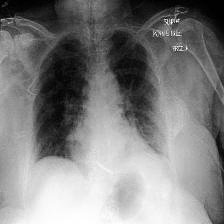} & \includegraphics[scale=0.4]{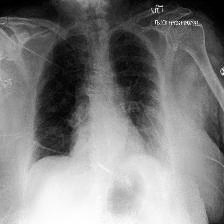} & \includegraphics[scale=0.4]{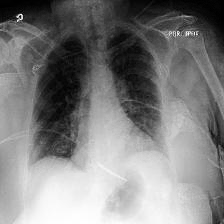} \\
    & \includegraphics[scale=0.4]{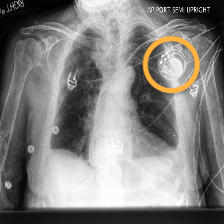}  & \includegraphics[scale=0.4]{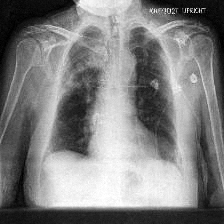} & \includegraphics[scale=0.4]{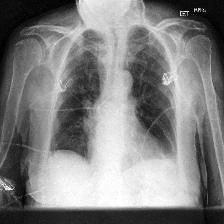} & \includegraphics[scale=0.4]{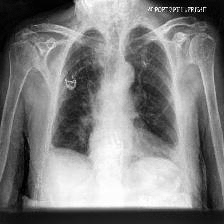} & \includegraphics[scale=0.4]{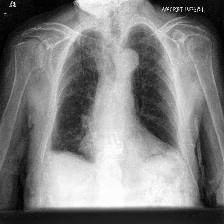} & \includegraphics[scale=0.4]{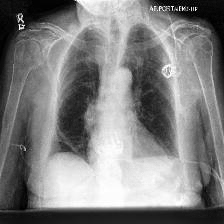} \\

    & \includegraphics[scale=0.4]{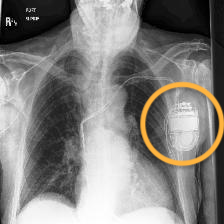}  & \includegraphics[scale=0.4]{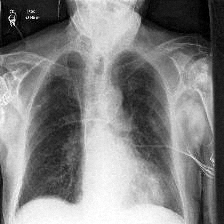} & \includegraphics[scale=0.4]{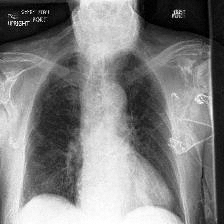} & \includegraphics[scale=0.4]{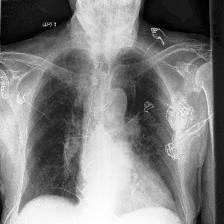} & \includegraphics[scale=0.4]{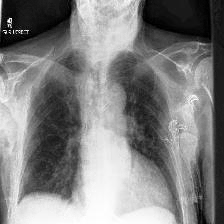} & \includegraphics[scale=0.4]{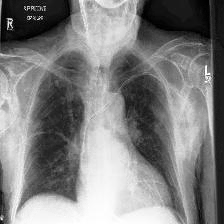} \\

    & \includegraphics[scale=0.4]{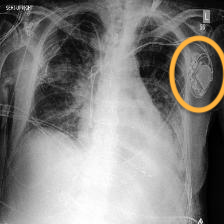}  & \includegraphics[scale=0.4]{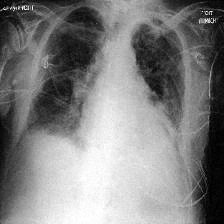} & \includegraphics[scale=0.4]{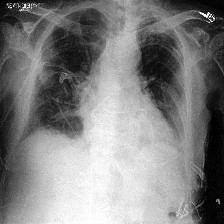} & \includegraphics[scale=0.4]{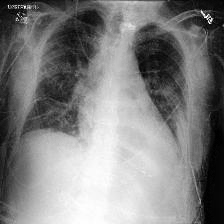} & \includegraphics[scale=0.4]{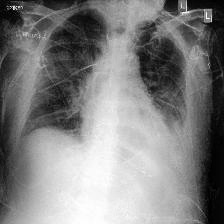} & \includegraphics[scale=0.4]{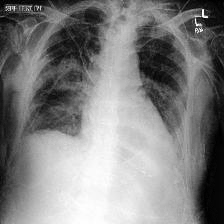} \\

     & \includegraphics[scale=0.4]{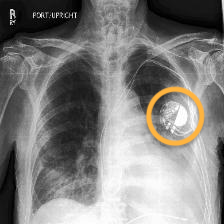}  & \includegraphics[scale=0.4]{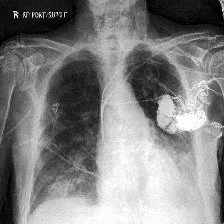} & \includegraphics[scale=0.4]{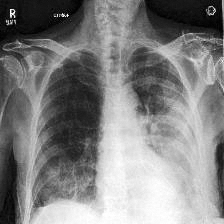} & \includegraphics[scale=0.4]{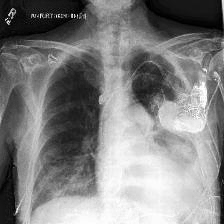} & \includegraphics[scale=0.4]{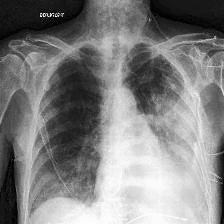} & \includegraphics[scale=0.4]{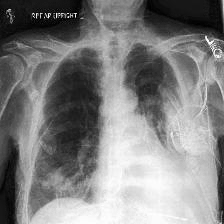} \\

     & \includegraphics[scale=0.4]{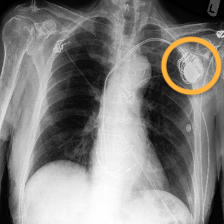}  & \includegraphics[scale=0.4]{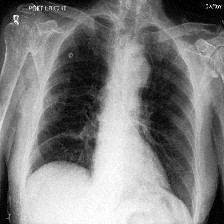} & \includegraphics[scale=0.4]{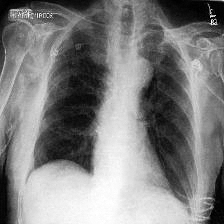} & \includegraphics[scale=0.4]{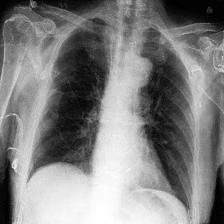} & \includegraphics[scale=0.4]{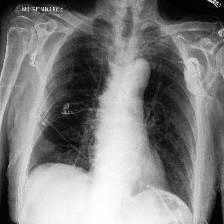} & \includegraphics[scale=0.4]{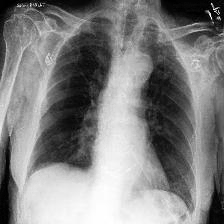} \\

\end{tabular}
}
\caption{\textbf{CheXpert -  Removing pacemakers.} The shortcuts are highlighted with orange circles or boxes in the original images.}
\label{fig:sup_che_visual}
\end{figure*}

\begin{figure*}
\centering
\scalebox{0.65}{
\begin{tabular}{ccccccc}
   & Original & DiME & \FastDiMEnonMask & \FastDiME & \FastDiMETwo & \FastDiMETwoPlus \\

   & \includegraphics[scale=0.4]{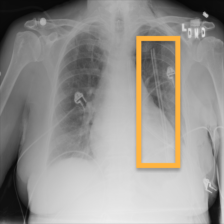}  & \includegraphics[scale=0.4]{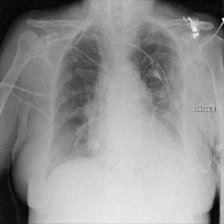} & \includegraphics[scale=0.4]{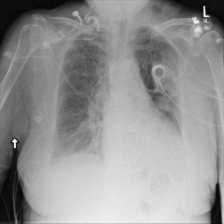} & \includegraphics[scale=0.4]{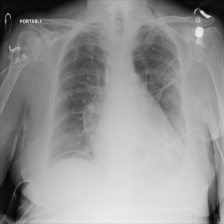} & \includegraphics[scale=0.4]{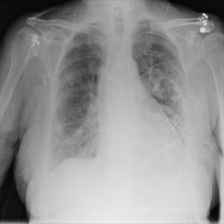} & \includegraphics[scale=0.4]{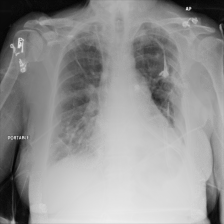} \\

   & \includegraphics[scale=0.4]{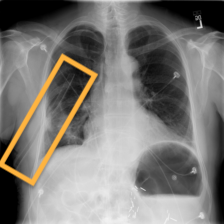}  & \includegraphics[scale=0.4]{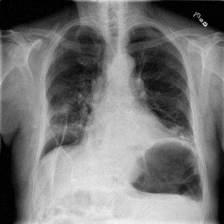} & \includegraphics[scale=0.4]{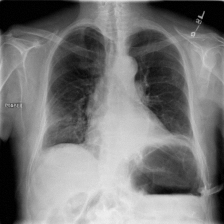} & \includegraphics[scale=0.4]{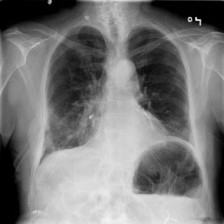} & \includegraphics[scale=0.4]{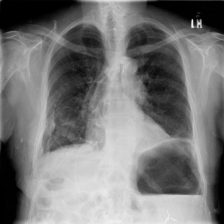} & \includegraphics[scale=0.4]{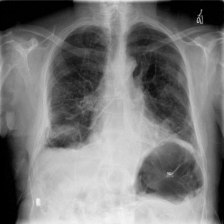} \\

   & \includegraphics[scale=0.4]{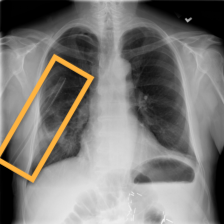}  & \includegraphics[scale=0.4]{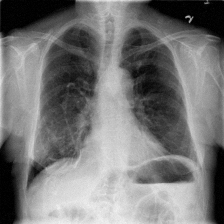} & \includegraphics[scale=0.4]{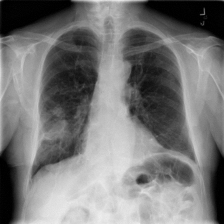} & \includegraphics[scale=0.4]{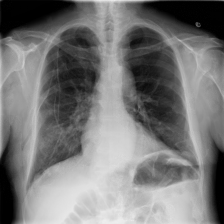} & \includegraphics[scale=0.4]{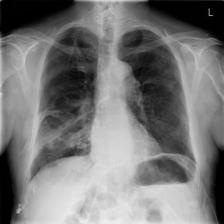} & \includegraphics[scale=0.4]{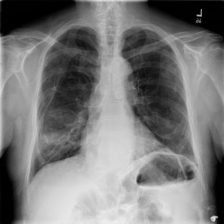} \\

   & \includegraphics[scale=0.4]{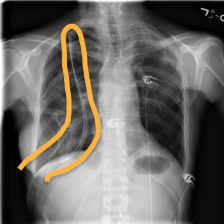}  & \includegraphics[scale=0.4]{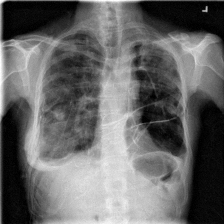} & \includegraphics[scale=0.4]{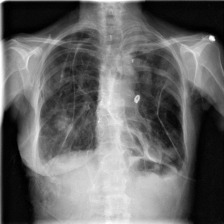} & \includegraphics[scale=0.4]{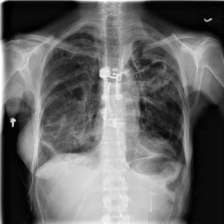} & \includegraphics[scale=0.4]{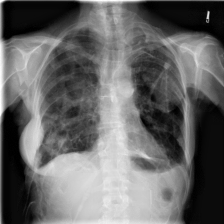} & \includegraphics[scale=0.4]{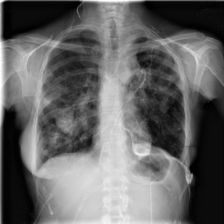} \\

    & \includegraphics[scale=0.4]{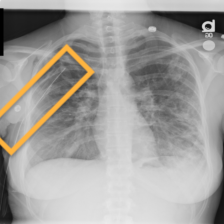}  & \includegraphics[scale=0.4]{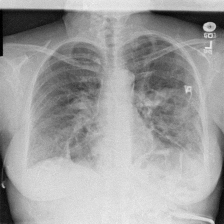} & \includegraphics[scale=0.4]{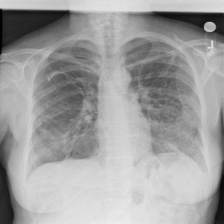} & \includegraphics[scale=0.4]{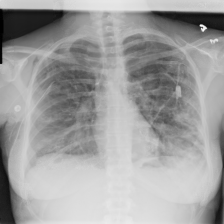} & \includegraphics[scale=0.4]{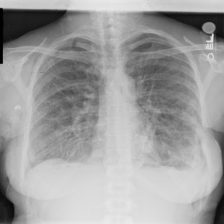} & \includegraphics[scale=0.4]{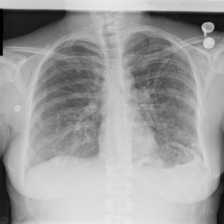} \\

    & \includegraphics[scale=0.4]{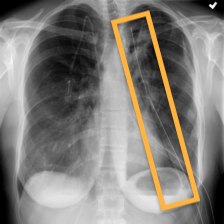}  & \includegraphics[scale=0.4]{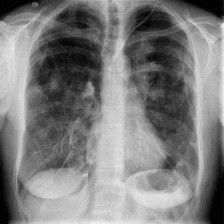} & \includegraphics[scale=0.4]{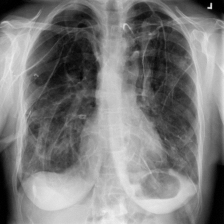} & \includegraphics[scale=0.4]{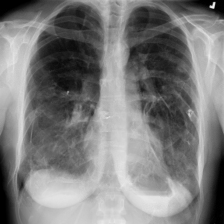} & \includegraphics[scale=0.4]{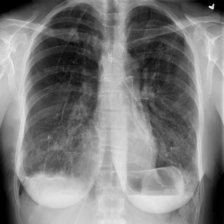} & \includegraphics[scale=0.4]{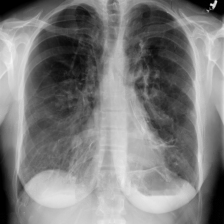} \\

    & \includegraphics[scale=0.4]{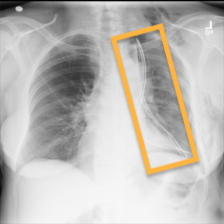}  & \includegraphics[scale=0.4]{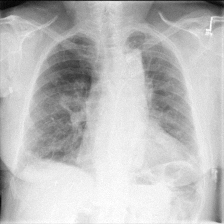} & \includegraphics[scale=0.4]{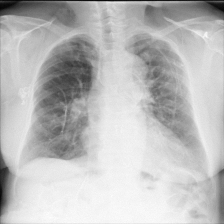} & \includegraphics[scale=0.4]{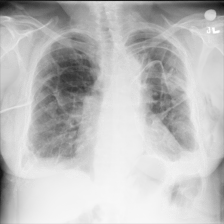} & \includegraphics[scale=0.4]{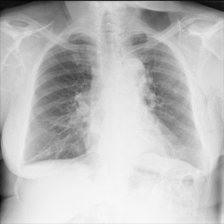} & \includegraphics[scale=0.4]{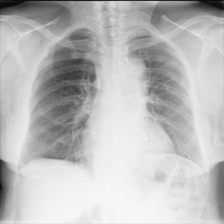} \\

    & \includegraphics[scale=0.4]{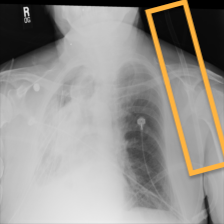}  & \includegraphics[scale=0.4]{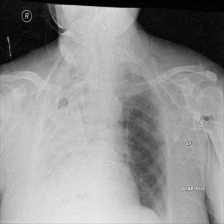} & \includegraphics[scale=0.4]{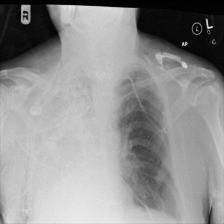} & \includegraphics[scale=0.4]{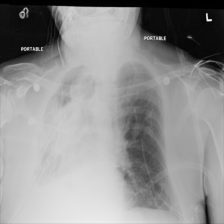} & \includegraphics[scale=0.4]{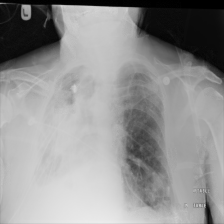} & \includegraphics[scale=0.4]{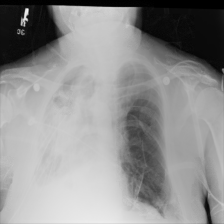} \\

\end{tabular}
}
\caption{\textbf{NIH - Removing chest drains.} The shortcuts are highlighted with orange circles or boxes in the original images.}
\label{fig:sup_nih_visual}
\end{figure*}

\begin{figure*}
\centering
\scalebox{0.65}{
\begin{tabular}{ccccccc}
   & Original & DiME & \FastDiMEnonMask & \FastDiME & \FastDiMETwo & \FastDiMETwoPlus \\

   & \includegraphics[scale=0.4]{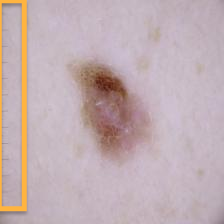}  & \includegraphics[scale=0.4]{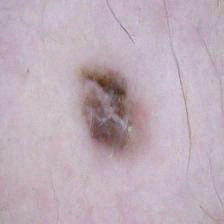} & \includegraphics[scale=0.4]{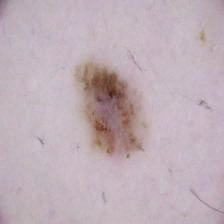} & \includegraphics[scale=0.4]{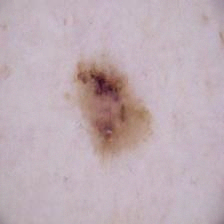} & \includegraphics[scale=0.4]{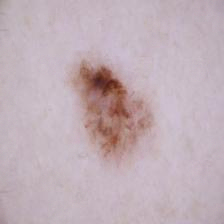} & \includegraphics[scale=0.4]{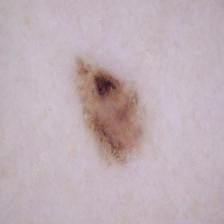} \\

   & \includegraphics[scale=0.4]{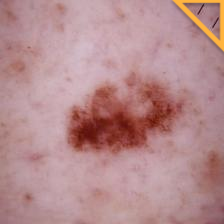}  & \includegraphics[scale=0.4]{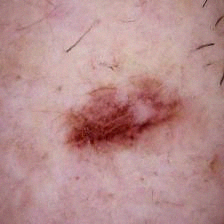} & \includegraphics[scale=0.4]{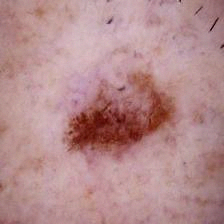} & \includegraphics[scale=0.4]{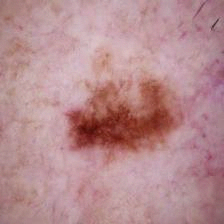} & \includegraphics[scale=0.4]{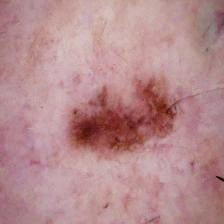} & \includegraphics[scale=0.4]{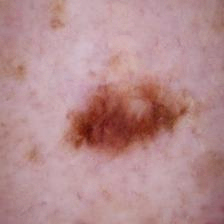} \\

   & \includegraphics[scale=0.4]{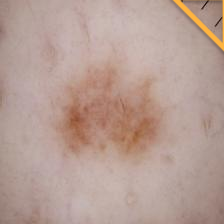}  & \includegraphics[scale=0.4]{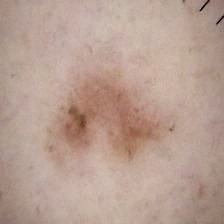} & \includegraphics[scale=0.4]{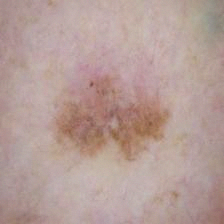} & \includegraphics[scale=0.4]{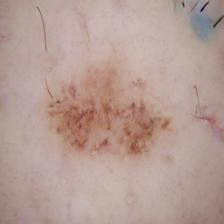} & \includegraphics[scale=0.4]{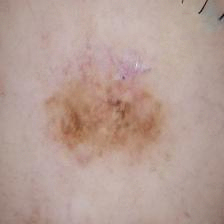} & \includegraphics[scale=0.4]{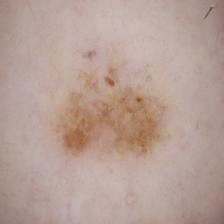} \\

   & \includegraphics[scale=0.4]{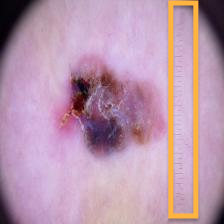}  & \includegraphics[scale=0.4]{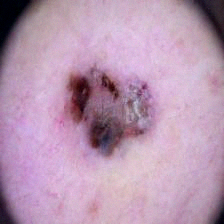} & \includegraphics[scale=0.4]{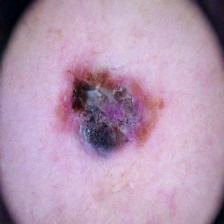} & \includegraphics[scale=0.4]{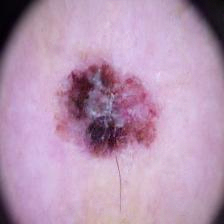} & \includegraphics[scale=0.4]{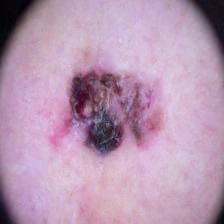} & \includegraphics[scale=0.4]{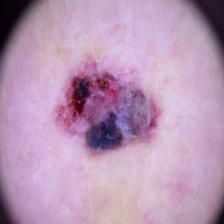} \\

   & \includegraphics[scale=0.4]{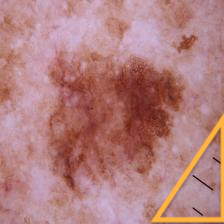}  & \includegraphics[scale=0.4]{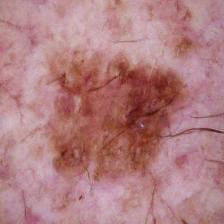} & \includegraphics[scale=0.4]{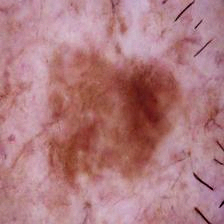} & \includegraphics[scale=0.4]{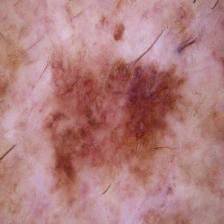} & \includegraphics[scale=0.4]{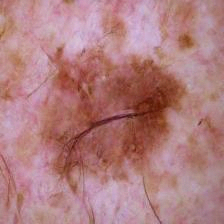} & \includegraphics[scale=0.4]{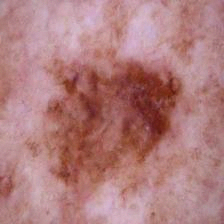} \\

    & \includegraphics[scale=0.4]{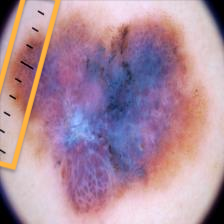}  & \includegraphics[scale=0.4]{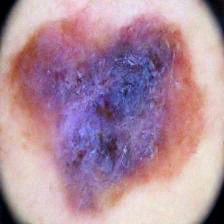} & \includegraphics[scale=0.4]{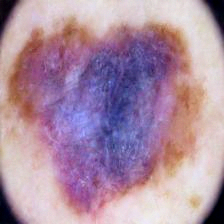} & \includegraphics[scale=0.4]{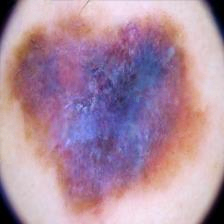} & \includegraphics[scale=0.4]{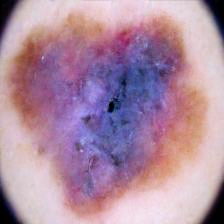} & \includegraphics[scale=0.4]{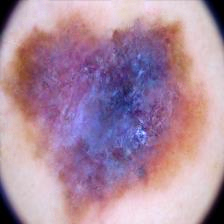} \\

    & \includegraphics[scale=0.4]{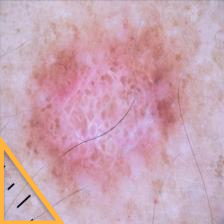}  & \includegraphics[scale=0.4]{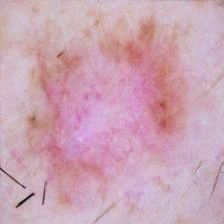} & \includegraphics[scale=0.4]{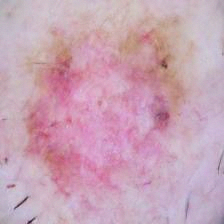} & \includegraphics[scale=0.4]{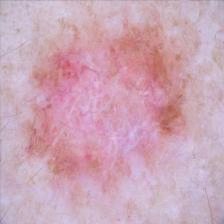} & \includegraphics[scale=0.4]{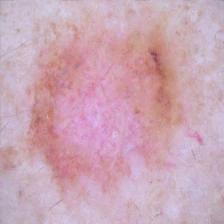} & \includegraphics[scale=0.4]{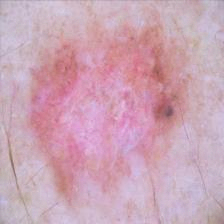} \\

    & \includegraphics[scale=0.4]{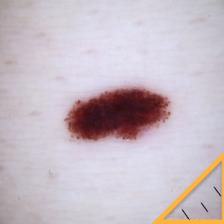}  & \includegraphics[scale=0.4]{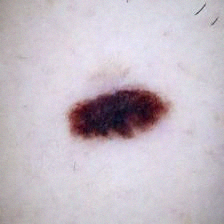} & \includegraphics[scale=0.4]{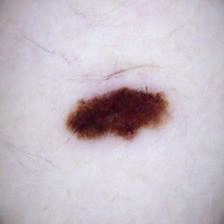} & \includegraphics[scale=0.4]{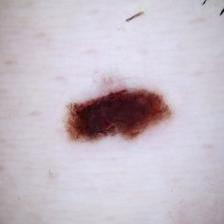} & \includegraphics[scale=0.4]{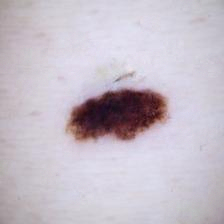} & \includegraphics[scale=0.4]{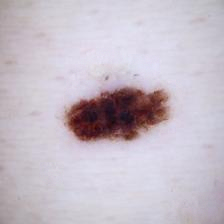} \\

\end{tabular}
}
\caption{\textbf{ISIC - Removing ruler markers.} The shortcuts are highlighted with orange circles or boxes in the original images.}
\label{fig:sup_isic_visual}
\end{figure*}

\begin{figure*}
\centering
\scalebox{0.7}{
\begin{tabular}{ccccccc}
   & Original & DiME & \FastDiMEnonMask & \FastDiME & \FastDiMETwo & \FastDiMETwoPlus \\

   & \includegraphics[width=30mm]{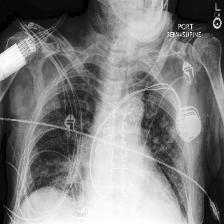}  & \includegraphics[width=30mm]{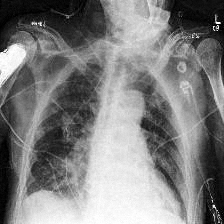} & \includegraphics[width=30mm]{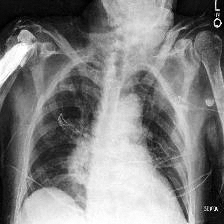} & \includegraphics[width=30mm]{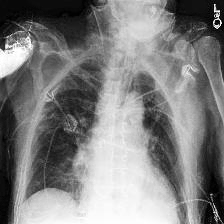} & \includegraphics[width=30mm]{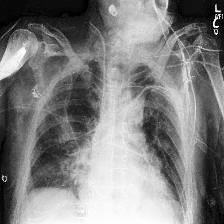} & \includegraphics[width=30mm]{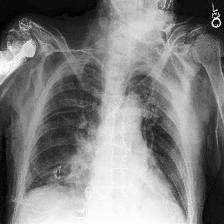} \\
   
    \rotatebox{90}{Difference Map}  
    &
    & \includegraphics[width=30mm]{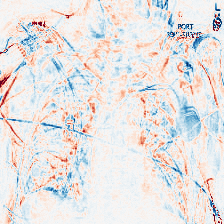} 
    & \includegraphics[width=30mm]{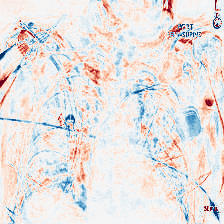} & \includegraphics[width=30mm]{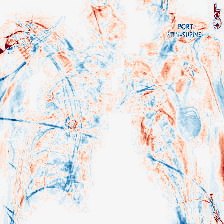} & \includegraphics[width=30mm]{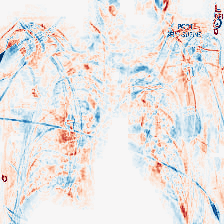} & \includegraphics[width=30mm]{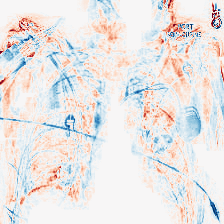} \\
    \cmidrule{2-7}
    & \includegraphics[width=30mm]{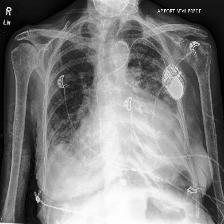}  & \includegraphics[width=30mm]{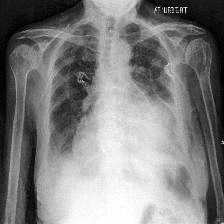} & \includegraphics[width=30mm]{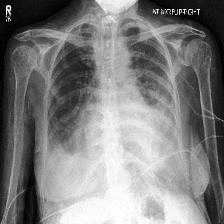} & \includegraphics[width=30mm]{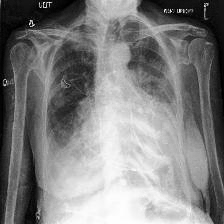} & \includegraphics[width=30mm]{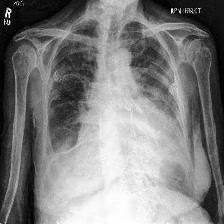} & \includegraphics[width=30mm]{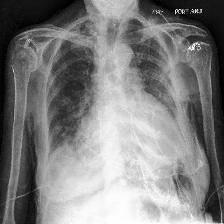} \\
   
    \rotatebox{90}{Difference Map}  
    &
    & \includegraphics[width=30mm]{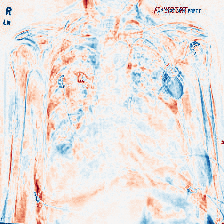} 
    & \includegraphics[width=30mm]{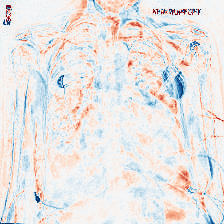} & \includegraphics[width=30mm]{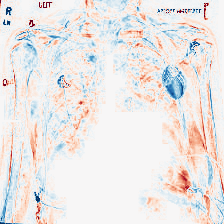} & \includegraphics[width=30mm]{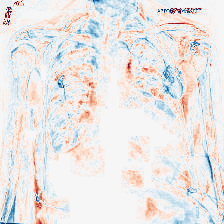} & \includegraphics[width=30mm]{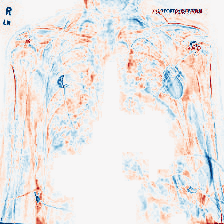} \\
    \cmidrule{2-7}
    & \includegraphics[width=30mm]{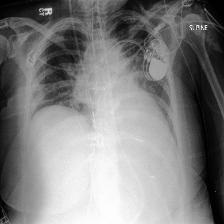}  & \includegraphics[width=30mm]{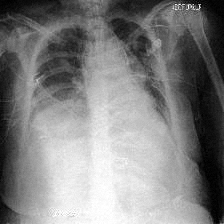} & \includegraphics[width=30mm]{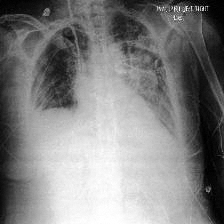} & \includegraphics[width=30mm]{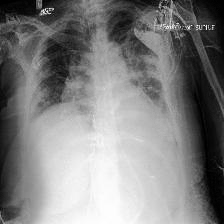} & \includegraphics[width=30mm]{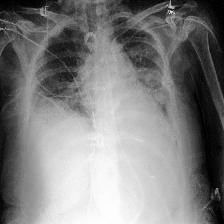} & \includegraphics[width=30mm]{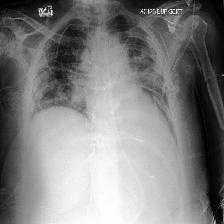} \\
   
    \rotatebox{90}{Difference Map}  
    &
    & \includegraphics[width=30mm]{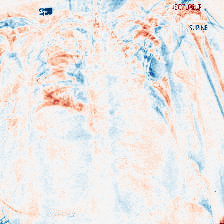} 
    & \includegraphics[width=30mm]{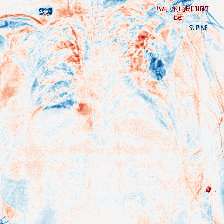} & \includegraphics[width=30mm]{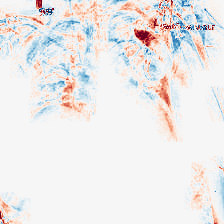} & \includegraphics[width=30mm]{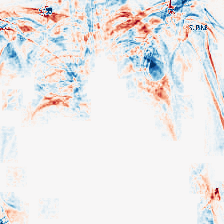} & \includegraphics[width=30mm]{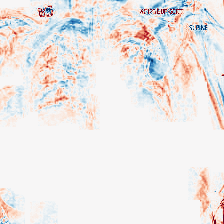} \\

\end{tabular}
}
\caption{\textbf{CheXpert - Removing pacemakers}. Examples together with corresponding difference maps between original and counterfactual images. Blue indicates a decrease in pixel values, and red indicates an increase in pixel values. }
\label{fig:sup_che_dm}
\end{figure*}

In this section, we show more qualitative results. Examples and comparisons against state-of-the-art methods on CelebA for `smile' and `age' attributes can be found in~\cref{fig:celebasmile} and~\cref{fig:celebaage}, respectively.
More visual examples of medical cases for shortcut counterfactuals can be found in~\cref{fig:sup_che_visual}, \cref{fig:sup_nih_visual}, and \cref{fig:sup_isic_visual}, for CheXpert samples (removing pacemaker), NIH samples (removing chest drain) and ISIC samples (removing ruler markers), respectively. We also provide examples of difference maps for CheXpert samples across different methods in~\cref{fig:sup_che_dm}.

\begin{table}[t]
    \centering
     \caption{\textbf{Ablation study on CheXpert.} The underlined hyperparameters are used in the main experiments. We highlight the \textbf{best} and \textit{second-best} performances.}
    \label{tab:ablationCHE}
    \scalebox{0.66}{\begin{tabular}{c|c|c|c|| c| c | c}
    \toprule
    Method & dilation & parameter $\tau_w$ & threshold & $L_1$ & MAD & FID \\
    \midrule
    \multirow{15}{*}{\FastDiME}
        & 5 & \multirow{5}{*}{\underline{80}} & \multirow{5}{*}{\underline{0.15}} & \textit{0.0484} & 0.7097 & \textit{52.2663}\\
        & 11 & & & 0.0698 & 0.7514 & 56.6191 \\
        &\underline{21} & && 0.0897 & 0.7554 & 61.4010\\
        & 31 & & & 0.0983 & \textit{0.7700} & 65.0084 \\
        & 55 & & & 0.1054 & 0.7689 & 67.1333\\
        \cmidrule{2-7}
        & \multirow{5}{*}{\underline{21}} & 10 & \multirow{5}{*}{\underline{0.15}}  & 0.0943 & 0.7128 & 66.5431\\
        & & 20 & & 0.0926 & 0.7114 & 65.4607 \\
        & & 40 & & 0.0898 & 0.6775 & 62.6441 \\
        & & \underline{80} & & 0.0897 & 0.7554 & 61.4010\\
        & & 160 & & 0.0955 & 0.7453 & 64.4503 \\
         \cmidrule{2-7}
        & \multirow{5}{*}{\underline{21}} & \multirow{5}{*}{\underline{80}} & 0.05 & 0.1077 & \textbf{0.7867 }& 66.7045 \\
        & & & 0.10 & 0.1025 & 0.7508 & 66.3789 \\
        & & & \underline{0.15} &  0.0897 & 0.7554 & 61.4010\\
        & & & 0.2 & 0.0710 & 0.7451 & 61.4010 \\
        & & & 0.3 & \textbf{0.0364} & 0.6465 & \textbf{39.2710} \\

        \midrule\midrule
    \FastDiMEnonMask & \ding{55} & \ding{55} &\ding{55}  &0.1082&0.7536&68.0411\\
    \bottomrule
    \end{tabular}
    }

\end{table}

\section{Ablation study}
\label{AblationStudy}

In this section, we quantitatively study the effect of our self-optimized masking scheme using CheXpert (see \cref{tab:ablationCHE}) CelebA (see \cref{tab:ablationCelebAage} and \cref{tab:ablationCelebAsmile}). For CelebA, we randomly select 4000 images and study the effect of our parameter $\tau_w$ as well as the impact of the dilation parameter. Moreover, we compare our self-optimized masking scheme with ACE~\cite{jeanneret2023adversarial}'s inpainting step which employs RePaint~\cite{lugmayr2022repaint} Our fully gradient-guided approach ensures the validity of counterfactuals while RePaint does not guarantee valid counterfactuals. For RePaint, we set the default hyperparameters for CelebA from ACE's~\cite{jeanneret2023adversarial} original implementation, i.e., $\tau = 5$ out of 50 respaced steps during inpainting. 
We also quantitatively evaluate GMD~\cite{karunratanakul2023guided} including our self-optimized masking scheme for the `smile' attribute in CelebA. Notably, GMD's noisy gradients lead to both lower quality and validity compared to our proposed method. 
Results for the `smile' and `age' attributes can be found in~\cref{tab:ablationCelebAsmile} and~\cref{tab:ablationCelebAage}, respectively.
For CheXpert, we also study the effect of the threshold. Increasing the threshold results in minimal changes to the original input image. Results can be found in \cref{tab:ablationCHE}.

\begin{table}[t]
    \centering
    \caption{\textbf{Ablation study on a subset of CelebA.} Results for the `smile' attribute. \FastDiMEnonMask{} refers to our method without our self-optimized masking scheme. We highlight the \textbf{best} performances.}
    \label{tab:ablationCelebAsmile}
    \scalebox{0.62}{\begin{tabular}{c|c|c||c|c|c|c|c|c}
    \toprule
    Method & dilation & parameter $\tau_w$ & FR & FID & FS & MNAC & CD & MAD \\
    \midrule
    \FastDiMEnonMask & \ding{55} & \ding{55} & \textbf{0.998} & 10.27 & 0.73 & 3.69 & 1.96 & \textbf{0.85} \\ \midrule
    \makecell{\FastDiMEnonMask \\ + RePaint}  & 15 & \ding{55} & 0.640 & 9.70 & 0.71 & 3.42 & 1.92 & 0.56 \\ \midrule

    GMD (/w Mask) & 15 & 30 & 0.881 & 8.45 & 0.75 & 3.37 & 2.37 & 0.71 \\ \midrule \midrule
    \multirow{6}{*}{\FastDiME}
        & & 30 & 0.990 & 8.19 & 0.76 & 3.09 & 1.98 & 0.82 \\
        & 5 & 45 & 0.988 & 9.76 & 0.74 & 3.27 & 2.08 & 0.82 \\
        & & 60 & 0.989 & 11.49 & 0.72 & 3.46 & \textbf{1.80} & 0.83 \\ \cmidrule{2-9}
        & & 30 & 0.997 & 8.68 & 0.73 & 3.43 & 2.01 & \textbf{0.85} \\
        & 15 & 45 & 0.997 & 8.61 & 0.73 & 3.41 & 2.10 & \textbf{0.85} \\
        & & 60 & \textbf{0.998} & 8.94 & 0.73 & 3.48 & 1.95 & \textbf{0.85} \\ \midrule
    \multirow{2}{*}{\FastDiMETwo}
        & 5 & \ding{55}/60 (fixed) & 0.996 & 7.48 & 0.77 & 3.01 & 1.94 & 0.83 \\ \cmidrule{2-9}
        & 15 & \ding{55}/60 (fixed) & \textbf{0.998} & 8.57 & 0.73 & 3.41 & 2.02 & \textbf{0.85} \\ \midrule
    \multirow{2}{*}{\FastDiMETwoPlus}
        & 5 & 30/60 (fixed) & 0.990 & \textbf{6.99} & \textbf{0.78} & \textbf{2.88} & 2.08 & 0.82 \\ \cmidrule{2-9}
        & 15 & 30/60 (fixed) & \textbf{0.998} & 8.64 & 0.73 & 3.43 & 1.93 & \textbf{0.85} \\

    \bottomrule
    \end{tabular}
    }

\end{table}

\begin{table}[t]
    \centering
    \caption{\textbf{Ablation study on a subset of CelebA.} Results for the `age' attribute. \FastDiMEnonMask{} refers to our method w/o our self-optimized masking scheme. We highlight the \textbf{best} performances.}
    \label{tab:ablationCelebAage}
    \scalebox{0.62}{\begin{tabular}{c|c|c||c|c|c|c|c|c}
    \toprule
    Method & dilation & parameter $\tau_w$ & FR & FID & FS & MNAC & CD & MAD \\
    \midrule
    \FastDiMEnonMask & \ding{55} & \ding{55} & 0.997 & 11.90 & 0.70 & 3.49 & 3.95 & \textbf{0.76} \\ \midrule
    \makecell{\FastDiMEnonMask \\ + RePaint}  & 15 & \ding{55} & 0.540 & 11.98 & 0.68 & 2.85 & 3.69 & 0.40 \\ \midrule\midrule
    
    \multirow{6}{*}{\FastDiME}
        & & 30 & 0.987 & 8.83 & 0.74 & 2.65 & 3.84 & 0.69 \\
        & 5 & 45 & 0.989 & 10.73 & 0.72 & 2.83 & 3.79 & 0.70 \\
        & & 60 & 0.995 & 12.70 & 0.70 & 3.11 & 3.72 & 0.73 \\ \cmidrule{2-9}
        & & 30 & 0.998 & 9.78 & 0.71 & 3.11 & 3.78 & 0.75 \\
        & 15 & 45 & 0.999 & 9.80 & 0.70 & 3.12 & 3.91 & 0.75 \\
        & & 60 & 0.998 & 9.93 & 0.70 & 3.13 & 3.81 & 0.75 \\ \midrule
    \multirow{2}{*}{\FastDiMETwo}
        & 5 & \ding{55}/60 (fixed) & 0.995 & 8.08 & 0.75 & 2.62 & 3.77 & 0.71 \\ \cmidrule{2-9}
        & 15 & \ding{55}/60 (fixed) & 0.998 & 9.59 & 0.70 & 3.13 & 3.69 & 0.75 \\ \midrule
    \multirow{2}{*}{\FastDiMETwoPlus}
        & 5 & 30/60 (fixed) & 0.988 & \textbf{7.49} & \textbf{0.76} & \textbf{2.43} & 3.99 & 0.69 \\ \cmidrule{2-9}
        & 15 & 30/60 (fixed) & \textbf{0.999} & 9.40 & 0.71 & 3.10 & \textbf{3.62} & 0.75 \\ 
    
    \bottomrule
    \end{tabular}
    }

\end{table}

\section{Denoised images for efficient gradient estimation}
\label{denoisedimages}
\begin{figure*}
\centering
\scalebox{0.75}{
\begin{tabular}{c|cccc}
    
   \textbf{\raisebox{4\height}{t=$\boldsymbol{\tau}$=60}} & \includegraphics[scale=0.6]{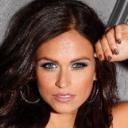}   & \includegraphics[scale=0.6]{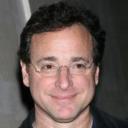} & \includegraphics[scale=0.6]{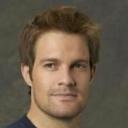} & \includegraphics[scale=0.6]{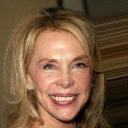} \\
   \textbf{\raisebox{4\height}{t=58}} & \includegraphics[scale=0.6]{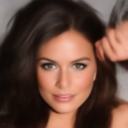}  & \includegraphics[scale=0.6]{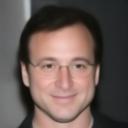} & \includegraphics[scale=0.6]{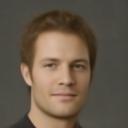} & \includegraphics[scale=0.6]{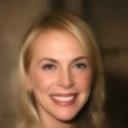}\\ 
   \textbf{\raisebox{4\height}{t=56}}& \includegraphics[scale=0.6]{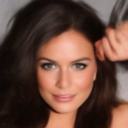}  & \includegraphics[scale=0.6]{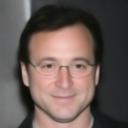} & \includegraphics[scale=0.6]{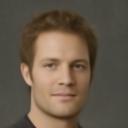} & \includegraphics[scale=0.6]{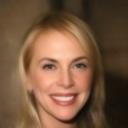}\\
   \textbf{\raisebox{4\height}{t=54}} & \includegraphics[scale=0.6]{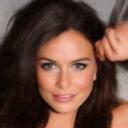}  & \includegraphics[scale=0.6]{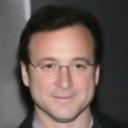}& \includegraphics[scale=0.6]{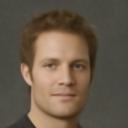} & \includegraphics[scale=0.6]{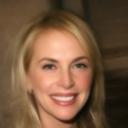}\\
   \textbf{\raisebox{4\height}{t=52}} & \includegraphics[scale=0.6]{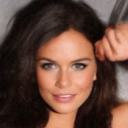}  & \includegraphics[scale=0.6]{figure/supp_denoised/smiletono_000066_0006.jpg}& \includegraphics[scale=0.6]{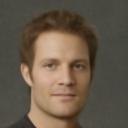} & \includegraphics[scale=0.6]{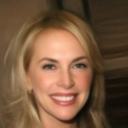}\\
   \textbf{\raisebox{4\height}{t=50}} & \includegraphics[scale=0.6]{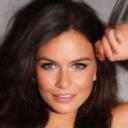}  & \includegraphics[scale=0.6]{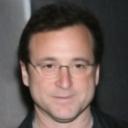}& \includegraphics[scale=0.6]{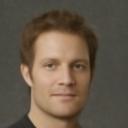} & \includegraphics[scale=0.6]{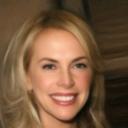}\\
   \textbf{\raisebox{4\height}{t=$\boldsymbol{\tau_w}$=30}}& \includegraphics[scale=0.6]{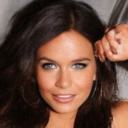}  & \includegraphics[scale=0.6]{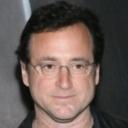}& \includegraphics[scale=0.6]{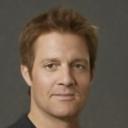} & \includegraphics[scale=0.6]{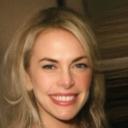}\\
   \textbf{\raisebox{4\height}{t=20}}& \includegraphics[scale=0.6]{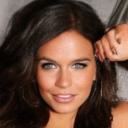}  & \includegraphics[scale=0.6]{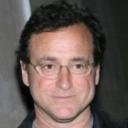}& \includegraphics[scale=0.6]{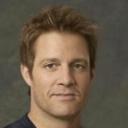} & \includegraphics[scale=0.6]{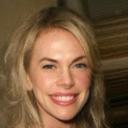}\\
   \textbf{\raisebox{4\height}{t=0}} & \includegraphics[scale=0.6]{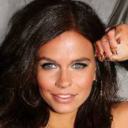}  & \includegraphics[scale=0.6]{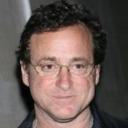}& \includegraphics[scale=0.6]{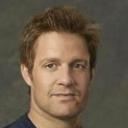} & \includegraphics[scale=0.6]{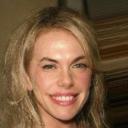} \\
   & no smile $\rightarrow$ smile &  smile $\rightarrow$ no smile  & young $\rightarrow$ old  & old $\rightarrow$ young  \\
\end{tabular}
}
\caption{\textbf{Counterfactual path}. The denoised image appears blurred at the start of the counterfactual generation process, but it progressively sharpens as the process advances toward completion. From time step $\tau_w = \frac{\tau}{2}$, denoised images are already sharp.} 
\label{fig:denoised_image}
\end{figure*}

\cref{fig:denoised_image} presents 4 visual examples from CelebA, illustrating the counterfactual path across early and late time steps. Initially, the denoised image appears blurred at the start of the counterfactual generation process, but it progressively sharpens as the process advances toward completion.

\section{Self-optimized masking}
\label{maskinpaint}
\begin{figure*}
\centering
\scalebox{0.7}{
\begin{tabular}{c|ccc|ccc}

   \textbf{\raisebox{4\height}{t=$\boldsymbol{\tau}$=60}} & \includegraphics[scale=0.6]{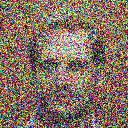}  & \includegraphics[scale=0.6]{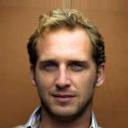} & & \includegraphics[scale=0.6]{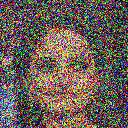} & \includegraphics[scale=0.6]{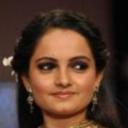} & \\
   
   \textbf{\raisebox{4\height}{t=55}} & \includegraphics[scale=0.6]{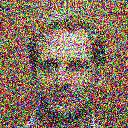}  & \includegraphics[scale=0.6]{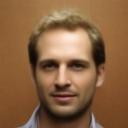} & & \includegraphics[scale=0.6]{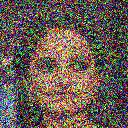} & \includegraphics[scale=0.6]{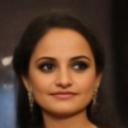} & \\

   \textbf{\raisebox{4\height}{t=$\boldsymbol{\tau_w}$=30}} & \includegraphics[scale=0.6]{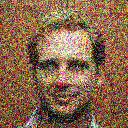}  & \includegraphics[scale=0.6]{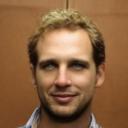} & \includegraphics[scale=0.6]{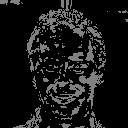} & \includegraphics[scale=0.6]{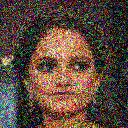} & \includegraphics[scale=0.6]{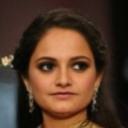} & \includegraphics[scale=0.6]{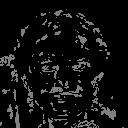} \\

   \textbf{\raisebox{4\height}{t=27}} & \includegraphics[scale=0.6]{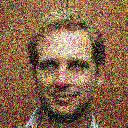}  & \includegraphics[scale=0.6]{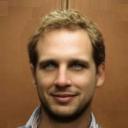} & \includegraphics[scale=0.6]{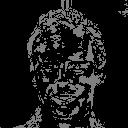} & \includegraphics[scale=0.6]{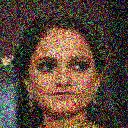} & \includegraphics[scale=0.6]{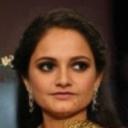} & \includegraphics[scale=0.6]{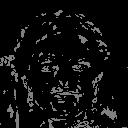} \\

   \textbf{\raisebox{4\height}{t=24}} & \includegraphics[scale=0.6]{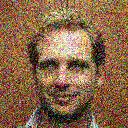}  & \includegraphics[scale=0.6]{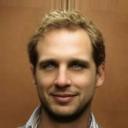} & \includegraphics[scale=0.6]{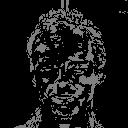} & \includegraphics[scale=0.6]{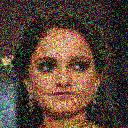} & \includegraphics[scale=0.6]{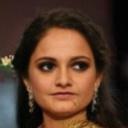} & \includegraphics[scale=0.6]{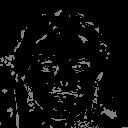} \\ 

   \textbf{\raisebox{4\height}{t=21}} & \includegraphics[scale=0.6]{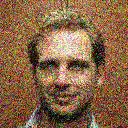}  & \includegraphics[scale=0.6]{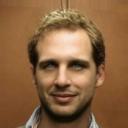} & \includegraphics[scale=0.6]{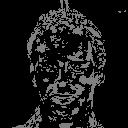} & \includegraphics[scale=0.6]{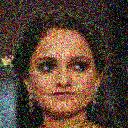} & \includegraphics[scale=0.6]{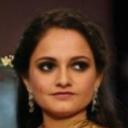} & \includegraphics[scale=0.6]{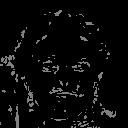} \\ 

   \textbf{\raisebox{4\height}{t=15}} & \includegraphics[scale=0.6]{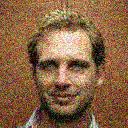}  & \includegraphics[scale=0.6]{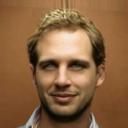} & \includegraphics[scale=0.6]{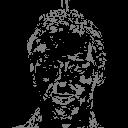} & \includegraphics[scale=0.6]{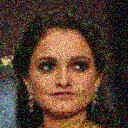} & \includegraphics[scale=0.6]{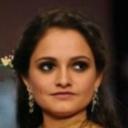} & \includegraphics[scale=0.6]{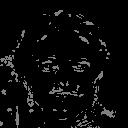} \\ 

   \textbf{\raisebox{4\height}{t=9}} & \includegraphics[scale=0.6]{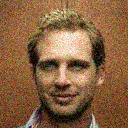}  & \includegraphics[scale=0.6]{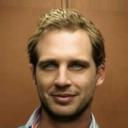} & \includegraphics[scale=0.6]{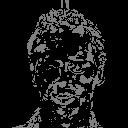} & \includegraphics[scale=0.6]{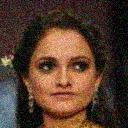} & \includegraphics[scale=0.6]{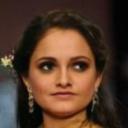} & \includegraphics[scale=0.6]{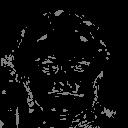} \\ 

   \textbf{\raisebox{4\height}{t=0}} & \includegraphics[scale=0.6]{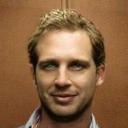}  & \includegraphics[scale=0.6]{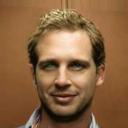} & \includegraphics[scale=0.6]{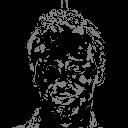} & \includegraphics[scale=0.6]{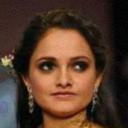} & \includegraphics[scale=0.6]{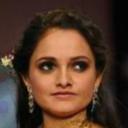} & \includegraphics[scale=0.6]{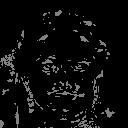} \\

   &  {\Large $x^c_t$ } & {\Large $\bar{x}^c_t$}  & {\Large $M_t$  }  &  {\Large $x^c_t$ } & {\Large $\bar{x}^c_t$} & {\Large $M_t$  } \\

\end{tabular}
}
\caption{\textbf{Self-optimized masking}. An example of how our method \FastDiME{} progresses across different time steps. We show the path toward the final counterfactual for two examples (left: no smile $\rightarrow$ smile, right: smile $\rightarrow$ no smile). We start applying our self-optimized masking scheme from $\tau_w = 30$. To better highlight the differences between steps we show the thresholded masks before applying dilation.}
\label{fig:maskinpaint}
\end{figure*}

\begin{figure*}
\centering
\scalebox{0.7}{
\begin{tabular}{ccccccc}
   Original & ACE $\ell_1$& DiME & \FastDiMEnonMask & \FastDiME & \FastDiMETwoPlus \\

  \includegraphics[scale=0.6]{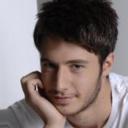}  & \includegraphics[scale=0.6]{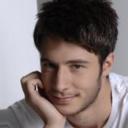} & \includegraphics[scale=0.6]{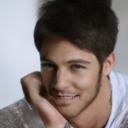} & \includegraphics[scale=0.6]{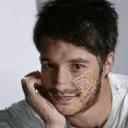} & \includegraphics[scale=0.6]{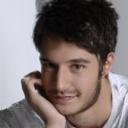} & \includegraphics[scale=0.6]{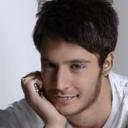} \\

  \includegraphics[scale=0.6]{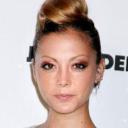}  & \includegraphics[scale=0.6]{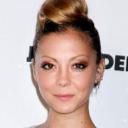} & \includegraphics[scale=0.6]{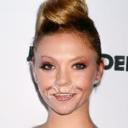} & \includegraphics[scale=0.6]{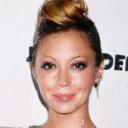} & \includegraphics[scale=0.6]{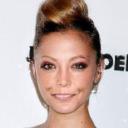} & \includegraphics[scale=0.6]{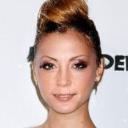} \\

   \includegraphics[scale=0.6]{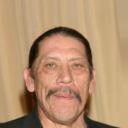}  & \includegraphics[scale=0.6]{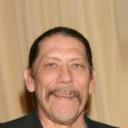} & \includegraphics[scale=0.6]{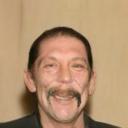} & \includegraphics[scale=0.6]{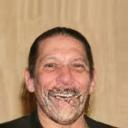} & \includegraphics[scale=0.6]{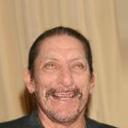} & \includegraphics[scale=0.6]{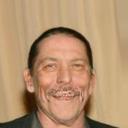} \\

   \includegraphics[scale=0.6]{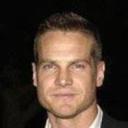}  & \includegraphics[scale=0.6]{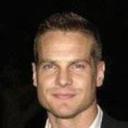} & \includegraphics[scale=0.6]{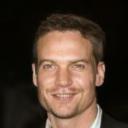} & \includegraphics[scale=0.6]{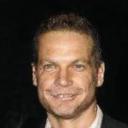} & \includegraphics[scale=0.6]{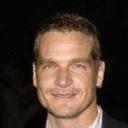} & \includegraphics[scale=0.6]{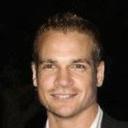} \\

\end{tabular}
}
\caption{\textbf{CelebA}. Examples for no smile $\rightarrow$ smile counterfactuals. We show examples of our method without the self-optimized masking scheme (FastDiME (w/o Mask)). Our self-optimized masking scheme helps to improve the results.}
\label{fig:celeba_star}
\end{figure*}

Initially, denoised images used for gradient estimation are blurred. We found both quantitatively (see \cref{AblationStudy}) and qualitatively (see \cref{denoisedimages}) that we need to start applying our self-optimized masking scheme as long as denoised images are not blurred, i.e, at $\tau_w = \frac{\tau}{2}$. An example of how our method \FastDiME{} progresses across different time steps is illustrated in \cref{fig:maskinpaint}, together with the thresholded masks $M_t$ used during our self-optimized masking scheme and the final absolute difference map between the counterfactual and the original image. To highlight its importance, we also provide visual examples of our method without the self-optimized masking scheme (\FastDiMEnonMask) in \cref{fig:celeba_star}.

% ------------- large figures --------------------

\end{document}